\documentclass[onecolumn]{article}

\usepackage{graphicx}
\usepackage{booktabs}
\usepackage{appendix}
\usepackage{adjustbox}
\usepackage{placeins}
\usepackage{amssymb}
\usepackage{enumitem}
\setlist[itemize]{label=\textbullet}
\usepackage{authblk}
\usepackage{etoolbox}
\usepackage{subcaption}

\usepackage[letterpaper, margin=1in]{geometry}
\usepackage{fancyhdr}
\newcommand{\boldparagraph}[1]{\paragraph{\normalfont{\textbf{#1}}}}

\usepackage[accsupp]{axessibility}  %
\usepackage[pagebackref,breaklinks, colorlinks, linkcolor=red, citecolor=blue]{hyperref}
\usepackage{cleveref}

\usepackage[numbers,sort&compress,merge,square]{natbib}
\usepackage{caption}
\crefname{figure}{Fig.}{Figs.}
\Crefname{figure}{Fig.}{Figs.}

\crefname{table}{Tab.}{Tabs.}
\Crefname{table}{Tab.}{Tabs.}

\crefname{section}{Sec.}{Secs.}
\Crefname{section}{Sec.}{Secs.}

\crefname{subsection}{Sec.}{Secs.}
\Crefname{subsection}{Sec.}{Secs.}

\title{A Closer Look at Benchmarking Self-Supervised Pre-training with Image Classification}
\date{}

\makeatletter
\renewcommand\paragraph{\@startsection{paragraph}{4}{\z@}%
  {3.25ex \@plus1ex \@minus.2ex}%
  {-1em}%
  {\normalfont\normalsize\itshape}}
\makeatother

\author[1]{Markus Marks\thanks{Equal contribution.}}
\author[1,2,3,4]{Manuel Knott$^*$}
\author[1]{Neehar Kondapaneni}
\author[1,5]{Elijah Cole}
\author[4]{\authorcr Thijs Defraeye}
\author[2,3,6]{Fernando Perez-Cruz}
\author[1]{Pietro Perona}

{
\scriptsize
\affil[1]{California Institute of Technology}
\affil[2]{ETH Zurich, Institute for Machine Learning, Department of Computer Science}
\affil[3]{Swiss Data Science Center, ETH Zurich and EPFL}
\affil[4]{Empa, Swiss Federal Laboratories for Materials Science and Technology}
\affil[5]{Altos Labs}
\affil[6]{Bank for International Settlements (BIS)}
}

\begin{document}

\maketitle
\setcounter{footnote}{0}

\begin{abstract}
\noindent
Self-supervised learning (SSL) is a machine learning approach where the data itself provides supervision, eliminating the need for external labels. 
The model is forced to learn about the data's inherent structure or context by solving a pretext task. 
With SSL, models can learn from abundant and cheap unlabeled data, significantly reducing the cost of training models where labels are expensive or inaccessible.
In Computer Vision, SSL is widely used as pre-training followed by a downstream task, such as supervised transfer, few-shot learning on smaller labeled data sets, and/or unsupervised clustering. Unfortunately, it is infeasible to evaluate SSL methods on all possible downstream tasks and objectively measure the quality of the learned representation. Instead, SSL methods are evaluated using in-domain evaluation protocols, such as fine-tuning, linear probing, and k-nearest neighbors (kNN).
However, it is not well understood how well these evaluation protocols estimate the representation quality of a pre-trained model for different downstream tasks under different conditions, such as dataset, metric, and model architecture.
In this work, we study how classification-based evaluation protocols for SSL correlate and how well they predict downstream performance on different dataset types. 
Our study includes eleven common image datasets and 26 models that were pre-trained with different SSL methods or have different model backbones.
We find that in-domain linear/kNN probing protocols are, on average, the best general predictors for out-of-domain performance. 
We further investigate the importance of batch normalization for the various protocols and evaluate how robust correlations are for different kinds of dataset domain shifts.
In addition, we challenge assumptions about the relationship between discriminative and generative self-supervised methods, finding that most of their performance differences can be explained by changes to model backbones. 

\end{abstract}

\renewcommand{\thefootnote}{\fnsymbol{footnote}}
\footnotetext[0]{Code available at: \url{https://github.com/manuelknott/ssl_eval_protocols}}
\renewcommand{\thefootnote}{\arabic{footnote}}
\setcounter{footnote}{0}

\section{Introduction}
\label{sec:introduction}

\begin{figure*}[ht!]
    \centering
    \includegraphics[width=\textwidth]{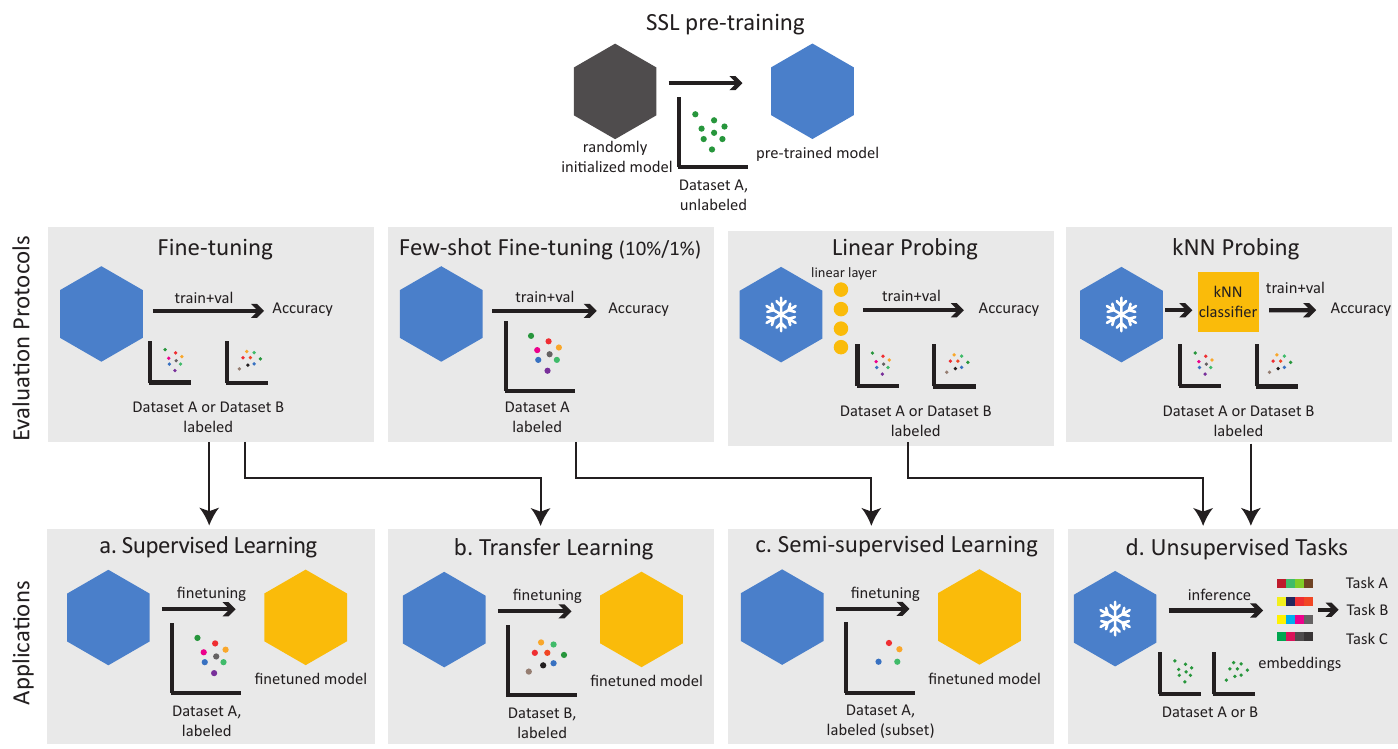}
    \caption{SSL application scenarios: We illustrate the following applications of self-supervised learning: 
    a) supervised learning (training and fine-tuning on the same dataset), 
    b) transfer learning (train on a large dataset and fine-tune the model on a ---usually smaller---domain dataset), 
    c) semi-supervised learning (train on a large unlabeled dataset and fine-tune on a small labeled subset of it), 
    d) unsupervised tasks (train on a dataset and run inference with the resulting model on any dataset to create embeddings that can be used for downstream tasks other than classification).
    Arrows between protocols and applications indicate a direct relationship. 
    }
    \label{fig:intro}
\end{figure*}

There has been a trend in machine learning where algorithmic improvements follow challenges posed through new datasets and evaluation metrics. 
How we evaluate new ML methods is therefore crucial, as the community may optimize for flawed \citep{locatello_challenging_2019} or misleading \citep{musgrave_metric_2020} metrics.
Self-supervised learning (SSL) is a promising path to advance machine learning using unlabeled data. 
It describes techniques that enable learning general image representations from abundant and cheap unlabeled data by solving pretext tasks \citep{balestriero_cookbook_2023}. 
Because of its effectiveness, SSL in computer vision has been used in a wide array of domains, ranging from animal behavior~\cite{sun_mabe22_2023}, retinal disease detection~\cite{zhou2023foundation}, computational histopathology~\cite{chen2023general} to remote sensing~\cite{wang2022self}.
SSL has proven more robust to data distribution shifts than supervised learning \citep{shi_how_2022}. 
Fundamentally, SSL methods aim to learn a general representation useful for {\em any} downstream task. 
What is ``downstream task performance'' and how should it be measured? We first consider the different applications of self-supervised pre-training. \Cref{fig:intro} (bottom) depicts the main practical applications where self-supervised pre-training is applied:

\begin{enumerate}
    \item[a.] \textit{Supervised Learning:} A model is pre-trained on dataset $A$ via SSL and then fine-tuned on the same dataset in a supervised way. This procedure can yield higher overall accuracies than supervised training from randomly initialized model weights \citep{he_masked_2021, bao_beit_2021}.

    \item[b.] \textit{Transfer Learning:} A model is pre-trained on dataset $A$ via SSL. The pre-trained backbone is then fine-tuned on a typically smaller labeled domain dataset $B$. This classical transfer learning paradigm can achieve better results with fewer data on small domain-specific data sets \citep{li_transfer_2020}. 
    SSL is usually a better starting point for transfer learning compared to supervised pre-training \citep{shi_how_2022}, as the latter is prone to overfit on features that are only useful for solving the initial supervised task \citep{jing_self-supervised_2019}.

    \item[c.] \textit{Semi-supervised Learning\footnote{The term \textit{semi-supervised learning} is commonly used to describe the ``self-supervised pretrain, supervised fine-tune on a subset'' paradigm (see e.g., \citep{chen_big_2020, chen_simple_2020, misra_self-supervised_2019}). We adopt this usage but acknowledge that, originally, semi-supervised learning refers to methods that utilize labeled and unlabeled data simultaneously rather than sequentially.}:} A model is pre-trained on an unlabelled dataset $A$ via SSL, followed by supervised fine-tuning on a small and labeled subset of the same dataset. This is particularly useful when data is cheap but labeling is expensive. 
    \citep{chen_big_2020, chen_simple_2020, grill_bootstrap_2020, zhou_ibot_2021}

    \item[d.] \textit{Unsupervised Tasks/Clustering}: A model is pre-trained on dataset $A$ via SSL. It is then used to generate embeddings at inference time. These embeddings can be used for various downstream tasks without further training the model \citep{pandarinath_inferring_2018, higgins_unsupervised_2021, sun_mabe22_2023}.
\end{enumerate}

\noindent

Evaluating the performance of SSL methods is challenging since there are endless ways to evaluate their learned representations, and exploring all of them is impossible. 
The community has developed several evaluation protocols to compare the representations' quality, resulting in {\em proxy metrics} for unobserved downstream tasks. 
Many of these protocols use the learned representation to solve classification tasks, for example, through linear probing, end-to-end fine-tuning, or by evaluating the embedding representation with a kNN classifier. 
The similarities and differences in the expressiveness of various protocols are understudied, which leads to an inconsistent evaluation and comparison of SSL methods (see \Cref{sec:survey}).
In this work, we show how reliably different protocols rank SSL methods w.r.t. to their performance on different downstream tasks. In detail, our contributions are as follows:

\begin{itemize}
    \item We survey existing papers on self-supervised learning methods for images and provide a structured summary of established evaluation protocols.
    \item We correlate in-domain (ID) and out-of-domain (OOD) top-1 and top-5 classification accuracies obtained from fine-tuning, linear probing, and kNN probing on 26 SSL-pretrained models. We show that linear/kNN probing protocols yield the proxy metrics that can, on average, best predict the {\em ranking} of SSL methods on eleven OOD datasets.
    \item We explore two kinds of domain shifts---categorical shift (either with coarse-grained or fine-grained features) and style shift---and find that in-domain proxy metrics vary in their predictiveness for each type of domain shift. 
    \item We compare generative and discriminative SSL protocols for ResNet and ViT backbones. We find that relative differences in linear probing and fine-tuning performance are more due to backbone architecture than the SSL family.
\end{itemize}

\section{Related Work}

\subsection{Self-Supervised Learning}

Self-supervised learning plays a crucial role in the recent success of natural language processing models \citep{qiu_pre-trained_2020, devlin_bert_2019} and computer vision \citep{chen_simple_2020, zhou_ibot_2021, caron_emerging_2021, he_masked_2021} and finds applications in tasks like speech recognition \citep{oord_representation_2018}, video classification \citep{feichtenhofer_masked_2022}, point cloud reconstruction \citep{yu_point-bert_2022} or behavioral analysis \citep{sun_mabe22_2023}. 
SSL relies on designing pretext tasks, forcing the model to learn a functional representation of the data without providing external labels \citep{balestriero_cookbook_2023}. 
Most SSL algorithms for images fall into one of two major categories: \textit{discriminative} and \textit{generative} methods \citep{liu_self-supervised_2021}. 

\boldparagraph{Discriminative methods.} Contrastive SSL methods for vision generate augmentations of samples and discriminate them from other samples in the data set \citep{chen_simple_2020, he_momentum_2019}. These methods rely on negative samples and, therefore, require large batch sizes.
A second line of work (\textit{self-distillation}) solely relies on positive samples~\citep{grill_bootstrap_2020, caron_unsupervised_2020}.
Yet another group of \textit{clustering-based} methods utilizes pseudo-labels based on k-means clustering in order to learn image representations \citep{caron_deep_2018, yan_clusterfit_2019}.

\boldparagraph{Generative methods.} Transformers~\citep{vaswani_attention_2017} are the current state-of-the-art deep-neural network architecture across many AI fields, bridging language and vision models. 
Inspired by pretext tasks for language transformer models, such as masking in BERT~\citep{devlin_bert_2019, he_masked_2021} recently introduced masked auto-encoding for images, an effective pre-training method, by which an image is split into patches, and about 70 percent of the patches are masked. 
Based on the remaining patches, the transformer will reconstruct the masked patches.
MaskFeat~\citep{wei_masked_2022} showed that the use of HOG features~\citep{dalal_histograms_2005} as reconstruction targets of masked patches is an effective pretext task.
Recent work combines masked image modeling with language-guided representations \citep{fang_eva_2022, hou_milan_2022}.
Another approach focuses on pixel-level reconstruction, alleviating the problem of missing foreground information that can occur with patch-based reconstruction approaches \cite{liu_pixmim_2023}.

\subsection{SSL Evaluation Protocols}

In general, self-supervised pre-training aims to learn useful representations across various downstream tasks. However, the quality of representations varies depending on the task. For example, some tasks may require representations invariant to certain transformations, while others may require representations preserving fine-grained details. For those reasons, designing evaluation protocols and associated metrics that capture all aspects is challenging. We conducted a literature survey on the different evaluation metrics used in SSL papers (\Cref{sec:survey}).
This section gives an overview of the most popular evaluation metrics. 
Typically, a study uses a set of a few metrics to evaluate the performance. 
This study mainly focuses on classification-based protocols, for which we identified four variations described in more detail below.
\Cref{fig:intro} illustrates their relationship to the previously mentioned use cases.

\boldparagraph{K-nearest neighbors (kNN).} kNN-classification is a way of probing the model, assuming that similar samples should have close Euclidean proximity in the latent space (see, e.g., \citep{caron_emerging_2021, caron_unsupervised_2020, zhou_ibot_2021, wu_unsupervised_2018}).
Compared to linear probing, kNN classifiers are fast and computationally light to deploy, often without an iterative learning setup \citep{caron_emerging_2021}. Since kNN requires no training, one could argue that this is the most direct and cheapest evaluation for representation learning.
However, clustering in high-dimensional spaces can be challenging \cite{assent_clustering_2012}. Another issue is that different dimensions do not necessarily have the same scale and might need to be normalized.

\boldparagraph{Linear probing.} 
In most cases, the classifier is implemented as a logistic regression model via a single fully-connected layer, usually referred to as \textit{Linear Probing} (see, e.g., \citep{caron_emerging_2021, grill_bootstrap_2020, chen_simple_2020, misra_self-supervised_2019, he_masked_2021, bao_beit_2021, chen_improved_2020,chen_generative_2020, dong_peco_2023, xie_simmim_2021, zhou_ibot_2021, goyal_self-supervised_2021}). 
The intuition here is that the learned representation is good if the dataset classes (the model was not trained on) are linearly separable. 
Besides linear and kNN probing, researchers sometimes use other shallow classifiers, e.g., Support Vector Machines \citep{caron_unsupervised_2020, wu_unsupervised_2018, doersch_unsupervised_2015, pathak_context_2016, zhang_split-brain_2016}. 

\boldparagraph{End-to-end fine-tuning.} Like linear probing, the end-to-end fine-tuning protocol replaces the last layer of a model with a linear classifier. In this setting, all model parameters are trained, allowing latent representations to adapt to the supervised task and/or a new data set (see, e.g., \citep{misra_self-supervised_2019, chen_simple_2020, chen_generative_2020, he_masked_2021, bao_beit_2021, chen_empirical_2021, zhou_ibot_2021, wei_masked_2022, hou_milan_2022}). Some papers use a partial fine-tuning protocol where only parts of the model are trained \citep{he_masked_2021, noroozi_unsupervised_2016, yosinski_how_2014}. 

\boldparagraph{Few-shot fine-tuning.} The few-shot learning protocol follows the same procedure as end-to-end fine-tuning but only uses a subset of the available training labels (typically 10\% or 1\%) (see, e.g., \citep{grill_bootstrap_2020, chen_simple_2020, caron_unsupervised_2020,zhou_ibot_2021, goyal_self-supervised_2021}), which makes evaluation significantly more efficient.\\

\noindent
Some common protocols that do not use classification are not included in the experimental part of this study but should be mentioned at this point. 
Our survey found that task transfer protocols, such as object detection \citep{misra_self-supervised_2019, he_masked_2021, chen_improved_2020, caron_unsupervised_2020, dong_peco_2023, zhou_ibot_2021}, semantic segmentation \citep{grill_bootstrap_2020, he_masked_2021, bao_beit_2021, misra_self-supervised_2019, zhou_ibot_2021}, depth estimation \citep{grill_bootstrap_2020}, copy detection \citep{caron_emerging_2021}, image retrieval \citep{caron_emerging_2021}, and super-resolution \citep{bao_beit_2021}, are frequently used to benchmark SSL methods.
Less commonly, unsupervised clustering, e.g., k-means, is used in the context of SSL evaluation (see, e.g., \citep{ zhou_ibot_2021, gansbeke_scan_2020}).

\subsection{Studies on SSL Evaluation Protocols}

While the community focuses on improving the capacity of SSL methods, evaluation protocols are seldom challenged. However, some studies can be used as references. 

\citet{kim2022broad} compared self-supervised and supervised pre-training for domain transfer. They evaluated fine-tuning accuracy on four downstream datasets for models pre-trained either supervised on ImageNet or SSL. By comparing four SSL methods, they found that supervised pre-training consistently outperformed SSL regarding OOD fine-tuning accuracy. As a shortcoming of their study, they mention the lack of possible combinations of different backbones with different SSL methods, which we address in our study.

\citet{yang_glue-x_2022} define an OOD benchmark for large language models. They find that distribution shifts between ID and OOD dominate OOD generalization results for language. They find discriminative models show a stronger linear correlation between ID and OOD performance than generative models. They find that linear probing shows relatively low ID and OOD accuracy, differing from findings in computer vision, where \citet{kumar_fine-tuning_2022} find that FT can do worse than LP for large distribution shifts.

\citet{newell_how_2020} find that the performance of an SSL algorithm in one setting might not translate to another. Moreover, they see that LP performance does not correlate with FT performance.
Linear transferability occurs when data from the same class in different domains are more related to each other than data from other classes in different domains~\citep{haochen_beyond_2022}. 

\citet{ibrahim_robustness_2022} measure the robustness of SOTA vision models, including SSL models, against distribution shifts w.r.t. factors of variation such as background, pose, etc., and find that the learning objective is more impactful for robustness than architecture.
Other studies have focused on understanding optimal SSL methods in the context of various metrics on ImageNet, such as fine-tuning accuracy, linear probing, and k-nearest neighbors \citep{ericsson_how_2021, gwilliam_beyond_2022}. 

\citet{miller_accuracy_2021} ask whether accuracy depends on in-domain to out-of-domain distributions shift, i.e. training on CIFAR-10~\citep{krizhevsky_learning_2009} and testing on CIFAR-10.1~\citep{recht_imagenet_2019}. They find that the linear trend between in-domain and out-of-domain performance holds across many but not all datasets. 

\citet{cole_when_2022} explore challenges in generalizing contrastive self-supervised learning beyond ImageNet, finding limitations with respect to data quantity, domain transfer, robustness, and fine-grained task performance.

Recently, \citet{goldblum_battle_2024} compared a wide range of architectural backbones and SSL setups on multiple datasets and downstream tasks. They focus on finding the backbone and method that generalizes best.
In contrast, our research focuses on which metric to use when developing SSL methods. 
Our findings challenge the wide use of fine-tuning as a metric \citep{he_masked_2021, feichtenhofer_masked_2022, wei_masked_2022}, as it does not strongly predict performance across different tasks and metrics. 

\citet{liu_self-supervised_2021} conducted a comparative study between discriminative and generative SSL methods among several domains (not limited to vision). They claim that contrastive learning methods---MoCo and SimCLR in particular---are effective if the downstream task is classification, while this is not obvious for many generation tasks. 

While some work has compared discriminative and generative models' influence on performance in vision~\citep{bao_beit_2021, wei_masked_2022, yu_vector-quantized_2021}, our study posits that the backbone of the model has more impact on performance than pre-training or pretext tasks. Specifically, we compare Vision Transformers (ViTs) with Residual Networks (ResNets), both indirectly and directly.

A recent study \citep{lee_rethinking_2023} presents a motivation akin to ours. However, the scope of our study is considerably larger as we compare more models (26 compared to 7), more OOD datasets (11 compared to 4), and a broader range of evaluation protocols, including three distinct fine-tuning protocols (100\%, 10\%, and 1\%).
Additionally, our study evaluates various model architectures, contrasting ResNets with Vision Transformers, and explores different types of domain shifts—categorical and style—by selecting transfer learning datasets.

\section{Experimental Setup}
\label{sec:experimental-setup}

\boldparagraph{Models and protocols.}
Our experiments are based on pre-trained models published by the original authors (if available) or replicas that achieve similar results to those reported in the original papers (see \Cref{sec:implementation_details} for sources of the pre-trained model weights). We use ResNet-50 and ViT-B16 backbones in this study. All models were pre-trained on ImageNet-1k~\citep{russakovsky_imagenet_2015}.  
We measure the accuracies of an SSL method on its training dataset (ImageNet) using five evaluation protocols: linear probing, kNN probing, and three variations of end-to-end fine-tuning with 100\%, 10\%, or 1\% of the available training data (see \Cref{fig:intro}, top). 
In addition, we compute kNN, linear probing, and fine-tuning (100\%) metrics on multiple OOD datasets for each SSL method. 

\boldparagraph{Correlation analysis.}
We spearman-correlate the results of the different protocols across 26 different SSL methods.
Linear and kNN probing are evaluated with and without normalizing the embedding. We found that normalization has no significant effect on some models and a large positive effect on others, especially those using masked image modeling. This aligns with findings from previous research \citep{lee_rethinking_2023}. All LP and kNN results reported in the main part of this paper use the normalized version (non-normalized results are reported in the supplementary materials).

\boldparagraph{OOD Datasets.}
For domain-shift analyses, we select our datasets following insights from previous work. It has been shown that the performance of current SSL models depends on the granularity of the dataset classes \citep{cole_when_2022}. We, therefore, choose datasets of different granularities in our study.
We chose Caltech-256 \citep{griffin_caltech_2022}, Pascal VOC 2012~\citep{everingham_pascal_2010}, and iNaturalist 2021 mini \citep{van_horn_benchmarking_2021} (``Family'' target) as representative datasets with coarse-grained classes.
In addition, we evaluate CUB \citep{wah_caltech-ucsd_2011} and two more variations of iNaturalist 2021 mini---with ``Genus'' or ``Species'' as target classes---as fine-grained datasets (see Appendix~\ref{appendix:inat} for details on how the iNaturalist datasets are constructed).
We also compare categorical domain shift and stylistic domain shift with respect to ImageNet. We group all previously mentioned datasets, excluding Pascal VOC, to create a group with no or few shared categories with ImageNet. This group is our categorical domain shift group. We use the ImageNet-D~\citep{rusak_imagenet-d_2022} dataset (ImageNet vocabulary but different styles) for stylistic domain shifts. 
A tabular overview of the dataset assignment described in this paragraph can be found in \Cref{tab:dataset-assignment}.

\boldparagraph{Hyperparameter selection.}
Usually, researchers sweep over a set of different hyperparameters during pre-training to find the best configuration for their method and evaluation protocols. This results in a variety of different hyperparameters for the same protocol. Therefore, it is very difficult to directly compare the reported metrics as they are confounded by the different choices of hyperparameters.
We decided to standardize our protocols by finding ``typical'' hyperparameter configurations for each of the protocols derived from the literature and use them for all models (see \Cref{sec:implementation_details} for implementation details). Consequently, the metrics we found in our experiments may deviate from the ones reported by the original authors. However, this standardization is crucial as our goal is not to benchmark the overall performance of different SSL methods but to correlate evaluation metrics under comparable conditions.

\boldparagraph{Robustness.}
In order to quantify variance introduced by random seeding, we calculate means and standard deviations for one model per protocol and dataset and repeat the same experiments for this selection three times (see \Cref{sec:appendix-errors}).

\section{Results}

\subsection{Which in-domain metric best predicts out-of-domain rankings on average?}

\begin{figure}[ht]
\centering

\begin{subfigure}[t]{0.45\textwidth} 
    \centering
        \includegraphics[width=\textwidth]{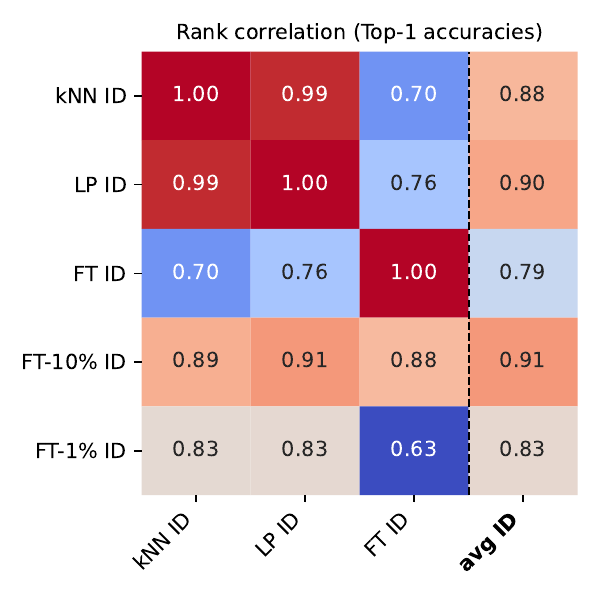}
\end{subfigure}
\hfill
\begin{subfigure}[t]{0.45\textwidth}
    \centering
        \includegraphics[width=\textwidth]{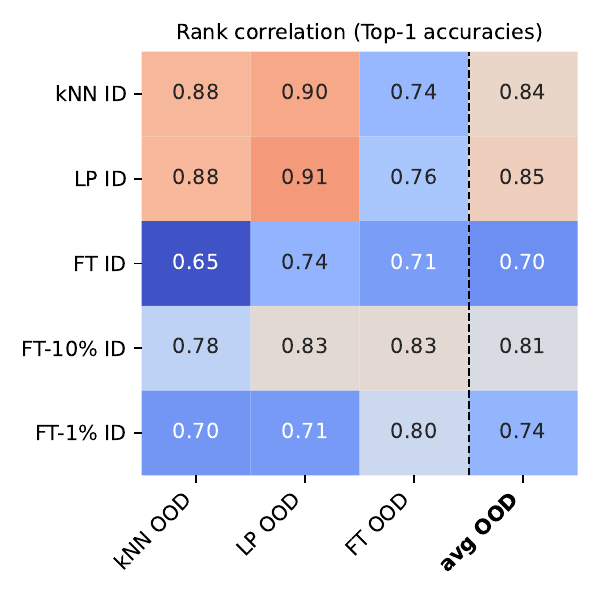}

\end{subfigure}

\caption{Comparing Spearman rank correlations of top-1 classification accuracies obtained by different evaluation protocols (kNN: k-nearest neighbors, LP: linear probing, FT: fine-tuning, FT-10\%: 10\%-fine-tuning, FT-1\%: 1\%-fine-tuning). In-domain (ID) refers to ImageNet-1k, which was also used for pre-training. Out-of-domain (OOD) metrics are averaged over eleven datasets as described in \Cref{sec:experimental-setup}.
In-domain metrics generally correlate highly (left panel), with fine-tuning having the weakest average correlation coefficient. When comparing ID with OOD protocols (right panel), correlation coefficients are visibly lower, indicating a domain-shift effect that impacts the absolute accuracy and the protocols' rank ordering (correlation).
A more verbose version of these matrices showing additional protocol variations (with and without feature normalization) is shown in \Cref{fig:corr-matrix-extended}.
}
\label{fig:corr-matrix}
\end{figure}

We begin our analysis by visualizing the rank correlations averaged across all models and datasets. In \Cref{fig:corr-matrix}, left panel, we see the ID metrics correlated against themselves. Expectedly, ID metrics generally correlate highly, with linear probing and 10\%-fine-tuning having the highest ($r=0.90$, $r=0.91$) and fine-tuning having the lowest average correlation coefficient ($r=0.79$).
Notably, Linear and kNN probing correlate almost perfectly ($r=0.99$) when features are normalized (\Cref{fig:corr-matrix}).
When comparing ID with OOD metrics, correlation coefficients are visibly lower, indicating that domain shifts affect both the absolute accuracy and also the ranking of different SSL representations. We will further investigate these effects in \Cref{sec:results_domainshift}.

Next, we target the question of whether an ID metric on training data is a good proxy metric for OOD use cases. We consider two use cases: (1) transfer learning, expressed by OOD fine-tuning accuracies averaged over multiple datasets, and (2) unsupervised representation learning, expressed by OOD kNN and/or linear probing accuracies averaged over multiple datasets.
We can observe that the probing protocols (kNN and linear) correlate most with themselves and each other when comparing ID and OOD accuracies.
Overall, linear probing is the best OOD predictor when averaged across metrics and datasets ($r=0.85$), followed closely by kNN  ($r=0.84$).
Interestingly, the two few-shot protocols appear to be the best predictors of model performance for out-of-distribution (OOD) fine-tuning. 
This correlation may arise from two primary factors: 
First, batch normalization on the embeddings is applied during the probing protocols but not for fine-tuning, which we discuss in more detail in \Cref{sec:emb-norm}.
Second, the transfer learning datasets in our study contain fewer training samples compared to ImageNet, resulting in a number of training steps during OOD fine-tuning that more closely resembles those in the few-shot in-distribution (ID) protocols rather than those used for fine-tuning on the full ImageNet dataset.

\paragraph{Summary: While probing protocols are the best OOD predictors on average, one should rely on few-shot fine-tuning (10\%) to predict the transfer learning capability for OOD fine-tuning.}

\subsection{How do protocols differ under different kinds of domain shift?}
\label{sec:results_domainshift}

\begin{figure}[htp]
\input{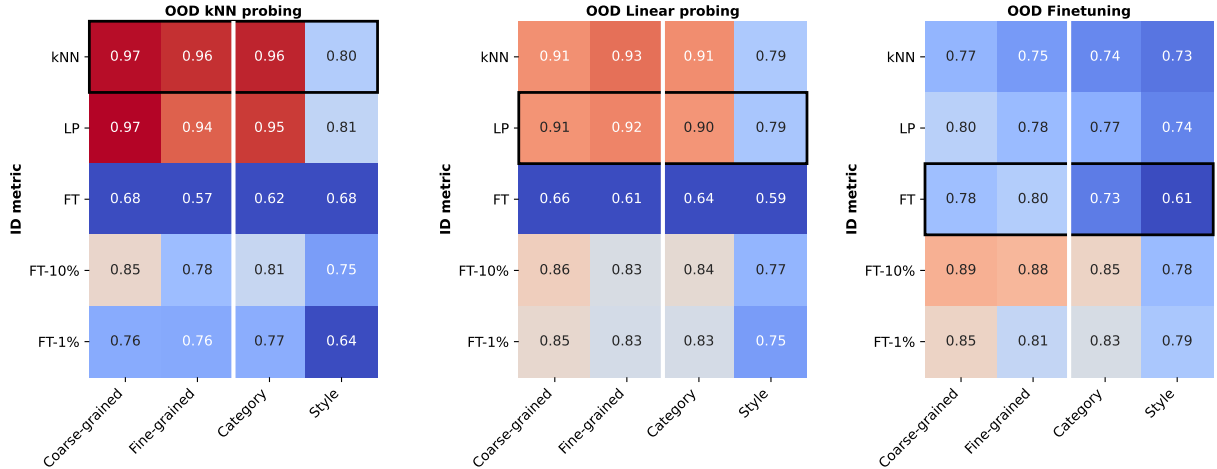} %
\caption{Spearman rank correlations of top-1 classification accuracies derived from in-domain and out-of-domain protocols under certain types of domain shift. We differentiate between fine-grained and coarse-grained categorical domain shifts (left half of each panel). Further, we compare categorical with stylistic domain-shift (right half of each panel).
Black rectangles highlight when the same ID and OOD evaluation protocol is used.
}
\label{fig:domainshift}
\end{figure}

\Cref{fig:domainshift} assesses the average rank correlations of top-1 accuracies under the four described domain shifts. In each panel, we explore the rank correlation of each in-domain metric against a single OOD metric averaged over domain shift grouped datasets. Generally, when comparing ID metrics with OOD metrics, we observe no notable difference between fine-grained and coarse-grained datasets but a significantly lower correlation for style-shift datasets.

\boldparagraph{OOD kNN.}
We can see that both ID probing protocols (kNN and linear) can reliably predict the ranking of OOD kNN accuracy for categorical domain shifts and less reliably for style-related domain shifts (left panel, top row). The correlation coefficient for categorical shift (kNN $r=0.96$, LP $r=0.95$) is notably higher than the equivalent for fine-tuning ($r=0.62$).

\boldparagraph{OOD LP.}
The general pattern is similar to OOD kNN probing, with ID kNN and ID LP being the strongest predictors and style shifts having a stronger impact than categorical shifts.
Few-shot fine-tuning protocols (1\% and 10\% correlate slightly more with OOD LP compared to OOD kNN. 

\boldparagraph{OOD FT.}
In \Cref{fig:corr-matrix}, ID FT is weakly correlated with OOD FT. In \Cref{fig:domainshift}, we see that FT rankings are less predictable with respect to shifts in both category ($r=0.73$) and style ($r=0.61$). Surprisingly, in-domain probing (ID LP, ID kNN) and few-shot fine-tuning (ID FT-10\% FT-1\%) protocols are better correlated to OOD FT across all types of domain shifts. As these protocols are significantly cheaper than full end-to-end fine-tuning (see \Cref{tab:computational_cost} for the estimated computational costs of each ID protocol in our experiments), they can be used as a proxy for the ranking of OOD fine-tuning performance.

\paragraph{Summary: 
The ranking of SSL methods is more robust for categorical and less for stylistic domain shifts under all protocols. There is no significant difference between fine-grained and coarse-grained categorical shifts.
}

\subsection{What is the effect of embedding normalization on different protocols?}
\label{sec:emb-norm}

Previous research has pointed out the importance of embedding normalization for linear \cite{he_masked_2021, lee_rethinking_2023} and kNN \cite{lee_rethinking_2023} probing.
Our experiments confirm that using batch normalization before the final classification layer in linear probing and z-score normalization for kNN probing can significantly increase accuracy (see \Cref{tab:results_all}). While the effect is strong for models with unscaled embedding representations (e.g., SimSiam and MaskFeat in our case), others (e.g., DINO) are neither positively nor negatively affected by normalization.

\begin{figure}[ht!]
    \centering
    \includegraphics[width=\textwidth]{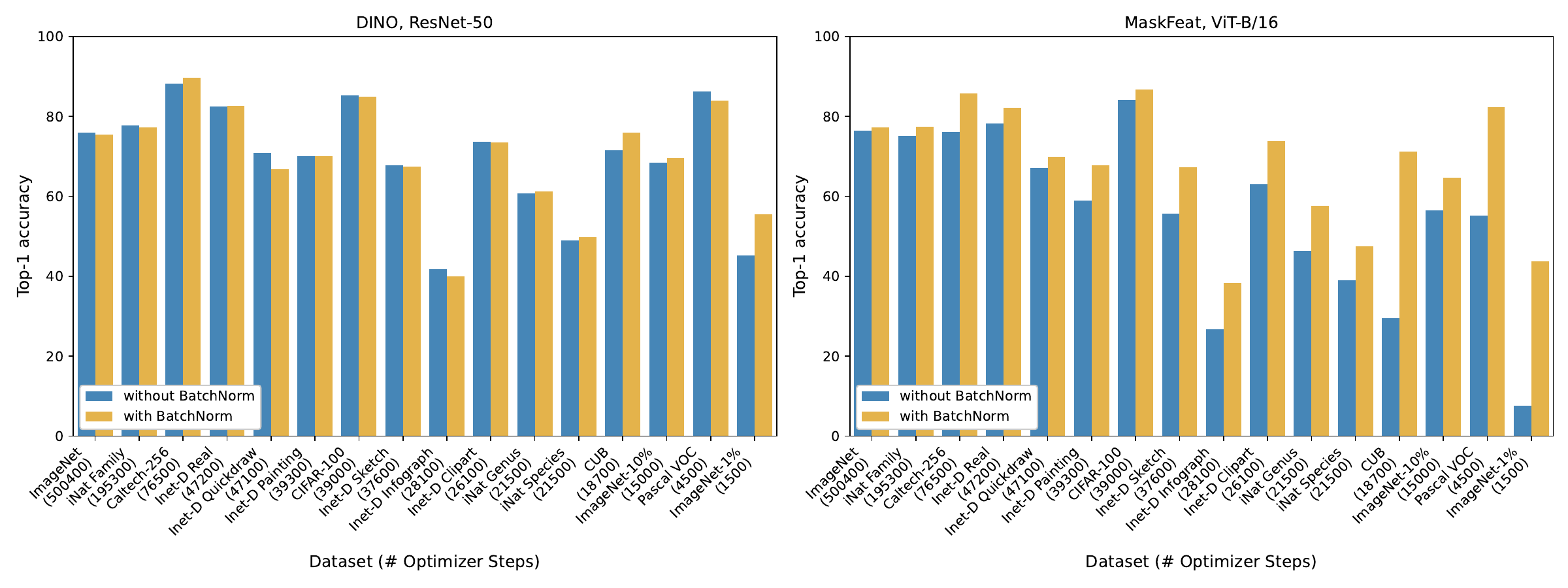}
    \caption{Fine-tuning accuracies with and without batch normalization for two exemplary models that appear to have scaled (left, DINO+ResNet-50) and unscaled (right, MaskFeat+ViT-B/16) embedding representations. The x-axes display all datasets included in this study and the number of optimizer steps derived from the dataset size, batch size, and total number of epochs.
    For MaskFeat, batch normalization has a significant effect when the number of optimizer steps is small and only a small effect when the number of steps is large, implying less-scaled features compared to DINO.}
    \label{fig:ft-bn}
\end{figure}

While batch normalization is common for linear probing, it is not established for fine-tuning protocols, presumably because feature scaling will be resolved during training when model weights are not frozen.
We challenge this assumption and claim that this is only true if a model is trained long enough (e.g., 100 epochs on full ImageNet) while fine-tuning on smaller datasets or with fewer epochs can yield significantly higher accuracies when BatchNorm is applied. \Cref{fig:ft-bn} displays fine-tuning accuracies with and without batch normalization for all the datasets included in our study, together with the total number of optimizer steps.

\paragraph{Summary: For models with unscaled features, batch normalization is critical for linear/kNN probing but also when fine-tuning on small datasets.}

\subsection{How do different SSL families and architectures perform under the various protocols?}
\label{sec:results_sslfamilies}

\begin{figure*}[tp]
    \centering
        \includegraphics[width=\textwidth]{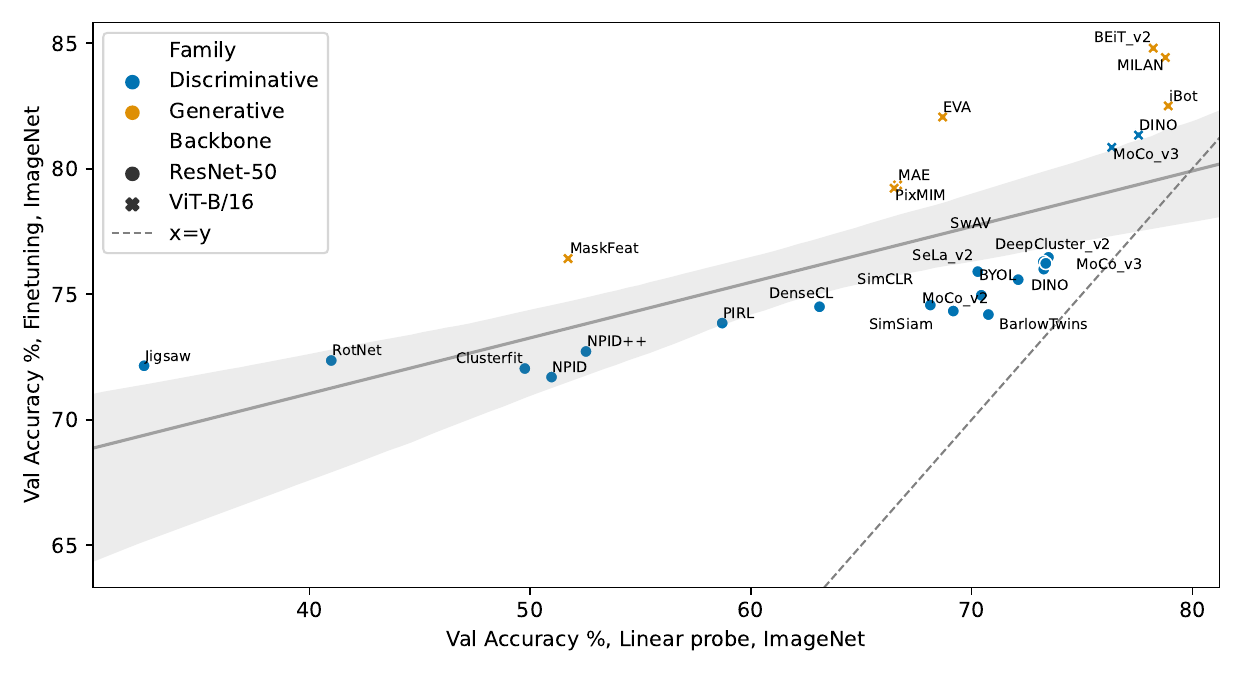}
    \caption{Scatter plot of the correlation of linear-probing and fine-tuning accuracies for ImageNet (in-domain). Each dot represents a model. The color codes for the model family, i.e., blue for discriminative and orange for generative models. Shapes indicate which backbones were used. The dotted line represents the equal error line; the solid line is a linear regression with a 90\% confidence interval.}
    \label{fig:lp_ft_correlation}
\end{figure*}

\begin{figure*}[ht!]
    \centering
    \includegraphics[width=\textwidth]{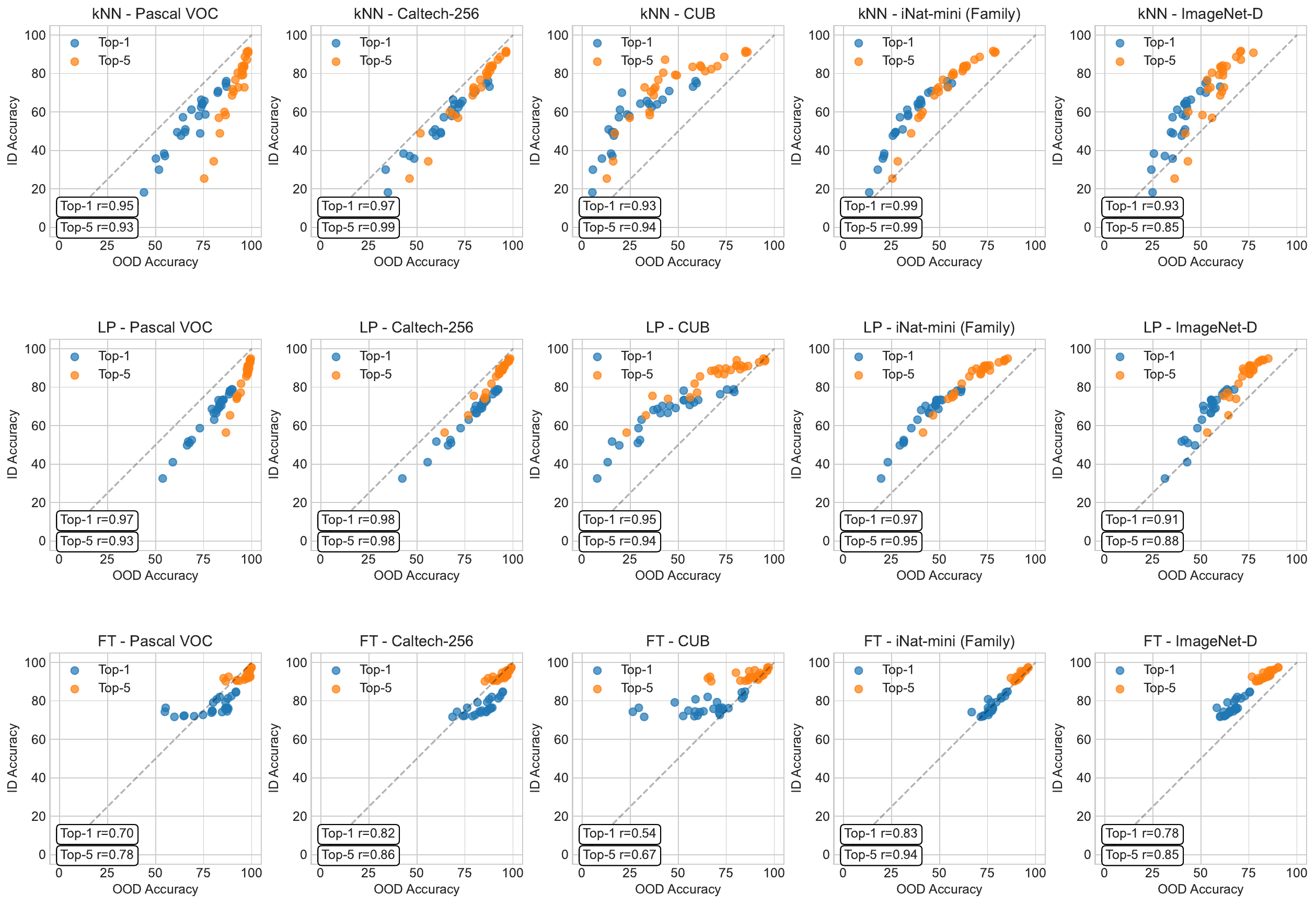}
    \caption{ID vs. OOD accuracy on different protocols and datasets. We compare both top-1 and top-5 classification accuracies. Correlation coefficients $r$ are calculated using Spearman's rank correlation. ImageNet-D accuracies are averaged across the six datasets. Individual ImageNet-D visualizations can be found in \Cref{fig:imagenet-d-panels}.}
    \label{fig:in_vs_out}
\end{figure*}

Previous work hypothesized that generative SSL methods achieve higher fine-tuning accuracies through expressive but non-linear features \citep{he_masked_2021}. Conversely, contrastive SSL methods achieve better linear probing performance through linearly separable features due to a discriminative loss function \citep{wei_masked_2022}. Following this hypothesis, recent studies have excluded linear probing altogether and have only used the fine-tuning protocol \citep{feichtenhofer_masked_2022}.

In \Cref{fig:lp_ft_correlation}, we plot the relation between each model's fine-tuning and linear probing performance on ImageNet.
We see that, indeed, the models with a generative loss (MaskFeat~\citep{wei_masked_2022}, MAE~\citep{he_masked_2021}, BEiT~v2~\citep{peng_beit_2022}, iBOT~\citep{zhou_ibot_2021}) have a larger performance gap between fine-tuning and linear probing performance on ImageNet than discriminate models. However, generative methods have been introduced in more recent publications, and ViT backbones are frequently used instead of CNNs. Could this cause the relative difference between fine-tuning and linear-probing performance? The figure shows that ViT backbones are all above the regression line, indicating a higher fine-tuning accuracy relative to their linear probing accuracy compared to other models. We can more directly assess this effect using two SSL methods that use both the ResNet-50 and ViT-B/16 backbones (DINO and MoCo-v3). For these models, we see that switching to a ViT backbone moves them from below the regression line to above. This suggests that the relatively higher fine-tuning accuracy is caused by a difference in backbone architectures rather than the SSL family,  which contrasts with previous hypotheses \citep{wei_masked_2022, he_masked_2021}).

\paragraph{Summary: Differences in linear probing performance between generative and discriminative models can be explained through different backbones rather than SSL methods.}

\subsection{How do rank correlations relate to absolute performance?}
\label{sec:rankings_vs_absolute_values}
Thus far, we have analyzed rank correlations between ID and OOD metrics. These correlations tell us how well the ranking of representations generated by an ID metric respects the ranking of the representations on the eleven OOD datasets we consider, and we have found several interesting trends. In \Cref{fig:in_vs_out}, we show how these trends manifest in terms of absolute performance, in which we visualize OOD accuracy against ID accuracy under each of our three metrics. Whether OOD accuracy will be higher or lower than ID accuracy depends on several factors, such as the representation quality and the similarity of the target dataset compared to ImageNet. For certain datasets (e.g., CUB and iNat-mini), OOD performances can be significantly worse even though the method ranking is very similar. On three datasets, Pascal VOC, Caltech-256, and CUB, we see that kNN and linear probing have a more linear relationship between ID and OOD accuracy than fine-tuning. On these datasets, we can see that several SSL methods can have almost the same ID fine-tuning accuracy but significantly different OOD fine-tuning accuracy. 

\paragraph{Summary: ID evaluation protocols can be robust proxies to estimate the \emph{ranking} of SSL methods, but not their absolute performance. 
}

\section{Discussion}

Self-supervision is a powerful way of leveraging unlabeled data for further downstream tasks.
Since the performance benchmark we choose will influence algorithmic development, it is crucial to evaluate SSL methods correctly for their intended purpose.
However, SSL evaluation is non-trivial since performance depends on the metric, the training dataset, the testing dataset, the computation required, and the downstream task.
We systematically investigate the performance of 26 SSL models on eleven datasets to evaluate which metric(s) should be used when benchmarking SSL models.

First, we find that linear and kNN probing accuracies are highly correlated when embedding normalization is applied. They can be used mostly interchangeably and are, on average, the best predictors for OOD metrics. 
Remarkably, we find that 10\%-fine-tuning on ImageNet is the strongest predictor for the ranking of SSL methods in OOD fine-tuning. This is particularly interesting for downstream users who are interested in using SSL pre-trained for transfer learning classification tasks.

Second, we find that linear/kNN probing is more robust to shifts in label granularity and (to some extent) style than fine-tuning protocols. 
When comparing ID accuracy against OOD accuracy directly, we see that several SSL methods can have equivalent ID fine-tuning results but much weaker OOD fine-tuning results~(\Cref{fig:in_vs_out}). 

Third, it was previously assumed that discriminative and generative SSL models differ in the type of representations they learn and that generative SSL methods result in powerful, but non-linear representations that require fine-tuning.
Using our comprehensive benchmark, we find that differences in performance may be attributed to the differences in the backbone used by different SSL methods.

Fourth, we investigate the importance of embedding normalization on several protocols. We confirm the findings of previous work \citep{lee_rethinking_2023} regarding the impact of batch normalization on batch protocols. In addition, we highlight the effects of batch normalization on end-to-end fine-tuning w.r.t. the dataset size.

\boldparagraph{Societal Impact.}
As the amount of data and applications for AI are growing, self-supervised learning plays an increasingly important role. SSL allows us to train models on unlabeled data and is necessary to reduce human annotation efforts and biases. Therefore, having an SSL evaluation metric that is predictive of various downstream tasks, i.e., of applications in the real world, is critical. In addition, SSL methods are more computationally expensive than supervised learning and, therefore, have a higher environmental impact. We must ensure that evaluation metrics accurately represent the utility of our methods, ensuring that time and resources spent on SSL development are not wasted.

\boldparagraph{Limitations.}
Our study has some limitations. While we cover several datasets and evaluation protocols, many more can still be considered. For example, SSL representations are commonly evaluated for other vision tasks like semantic segmentation, object detection, or depth estimation. Evaluating all of these is very costly and, therefore, beyond the scope of this study. 
Another limitation is categorizing each dataset as ID or OOD in a binary way. In future work, one could try to quantify dataset dissimilarity and use this as a proxy for how far out of distribution a dataset is.
Finally, future work should find theoretical grounding for our findings with respect to the interplay between SSL method, backbone, training dataset, and type of domain shift, similar to previous works~\citep{cabannes_ssl_2023}.

\section*{Acknowledgments}
Manuel Knott was supported by an ETH Zurich Doc.Mobility Fellowship. Pietro Perona and Markus Marks were supported by the National Institutes of Health (NIH R01 MH123612A) and the Caltech Chen Institute (Neuroscience Research Grant Award). Pietro Perona, Neehar Kondapaneni, and Markus Marks were supported by the Simons Foundation (NC-GB-CULM-00002953-02).

\bibliographystyle{unsrtnat}
\bibliography{references}

\clearpage
\setcounter{figure}{0}    
\setcounter{table}{0} 
\renewcommand\thefigure{S.\arabic{figure}} 
\renewcommand\thetable{S.\arabic{table}}
\appendix 
\begin{center}
    \begin{minipage}{\textwidth}
        \centering
        \fontsize{14}{18}\selectfont
        Supplementary Materials
    \end{minipage}
\end{center}
\vspace{6pt}
\section{Protocol Survey}
\label{sec:survey}

\cref{tab:survey}  shows the survey on classification-based SSL evaluation protocols we conducted at the beginning of this study. We decided to include the most protocols and to exclude very rarely used ones, i.e., partial fine-tuning and probing with a Support Vector Machine. Note that many papers additionally evaluate task transfer learning (e.g., semantic segmentation, object detection, image retrieval, or video classification), which is not covered in this survey.

\begin{table*}[ht!]
    \centering
    \small
    \caption{Survey on classification-based evaluation protocols for self-supervised learning in a selection of papers. A \checkmark denotes that the protocol was used in the mentioned paper. knn: k-nearest neighbors probing, LP: linear probing, FT: fine-tuning, FSFT: few-shot fine-tuning (1\% or 10\%), PFT: partial fine-tuning, SVM: SVM probing, CM: clustering metrics.
    Note that many papers additionally evaluate task transfer learning (typically on object detection or semantic segmentation tasks) which is not covered in this table.
    }
    \label{tab:survey}
    \begin{tabular}{lcccccccccc}
    \toprule
    & & \multicolumn{6}{c}{In domain} & \multicolumn{3}{c}{Out of domain} \\
    \cmidrule(lr){3-8} \cmidrule(lr){9-11} 
         Method & Year & knn & LP & FT & FSFT & PFT & SVM & knn  & LP & FT  \\
    \midrule
        Jigsaw \cite{noroozi_unsupervised_2016} & 2016 &  & & \checkmark  &  &  &  &  &  & \checkmark \\  
        npid \cite{wu_unsupervised_2018} & 2018 & \checkmark & & \checkmark & \checkmark & & \checkmark & & & \\
        PIRL \cite{misra_self-supervised_2019} & 2019  &  & \checkmark & \checkmark & \checkmark &  &  &  & \checkmark & \checkmark \\    
        BYOL \cite{grill_bootstrap_2020} & 2020        &  & \checkmark &  & \checkmark &  &  &  & \checkmark & \checkmark \\  
        SimCLR \cite{chen_simple_2020} & 2020          &  & \checkmark & \checkmark  & \checkmark &  &  &  & \checkmark & \checkmark \\
        MoCo v2 \cite{chen_improved_2020} & 2020       &  & \checkmark &  &  &  &  &  &  &  \\
        ImageGPT \cite{chen_generative_2020} & 2020    &  & \checkmark & \checkmark &  &  &  &  & \checkmark  & \checkmark  \\
        SwAV \cite{caron_unsupervised_2020} & 2020     & \checkmark &  & \checkmark & \checkmark &  & \checkmark &  & \checkmark &  \\
        SimSiam \cite{chen_exploring_2020} & 2020      & \checkmark & \checkmark &  &  &  &  &  &  &  \\
        DINO \cite{caron_emerging_2021} & 2021         & \checkmark & \checkmark & \checkmark & \checkmark &  &  &  & \checkmark \\
        MAE \cite{he_masked_2021}  &  2021             &  & \checkmark & \checkmark &  & \checkmark &  &  &  & \checkmark \\
        BEiT \cite{bao_beit_2021} & 2021               &  & \checkmark & \checkmark &  &  &  &  &  & \checkmark \\
        MoCo v3 \cite{chen_empirical_2021} & 2021      & \checkmark & \checkmark & \checkmark &  &  &  &  &  &  \\
        SimMIM \cite{xie_simmim_2021} & 2021           &  & \checkmark & \checkmark &  &  &  &  &  &  \\
        iBOT \cite{zhou_ibot_2021} & 2021              & \checkmark & \checkmark & \checkmark & \checkmark &  &  &  & & \checkmark \\
        MaskFeat \cite{wei_masked_2022} & 2021         &  &  & \checkmark &  &  &  &  &  &  \\
        SEER \cite{goyal_self-supervised_2021}  & 2021 &  & \checkmark & \checkmark & \checkmark &  &  &  & \checkmark & \checkmark \\
        Barlow Twins \cite{zbontar_barlow_2021} & 2021 &  & \checkmark &  & \checkmark &  &  &  & \checkmark &  \\
        data2vec \cite{baevski_data2vec_2022} & 2022   &  &  & \checkmark &  &  &  &  &  &  \\
        MILAN \cite{hou_milan_2022} & 2022 & & \checkmark & \checkmark & & & & & & \\
        PeCo \cite{dong_peco_2023} & 2023              &  & \checkmark & \checkmark &  &  &  &  &  &  \\
       PixMIM \cite{liu_pixmim_2023} & 2023 & &  \checkmark & \checkmark & \checkmark & & & & \\
        CAE \cite{chen_context_2024}            & 2024 &  & \checkmark & \checkmark &  &  &  &  &  & \checkmark \\
    \bottomrule
    \end{tabular}
    \end{table*}

\FloatBarrier
\section{Full Result Tables}
\label{sec:appendix-all-results}
(next page)
\begin{table*}[ht!]
    \centering
    \caption{
    {\em Top-1} classification accuracies (in \%). LP: Linear Probing, FT: End-to-end Fine-tuning, FT 10\%/1\%: Few-Shot Fine-tuning on 10\%/1\% of the training data, kNN: k-nearest neighbors probing.
    The superscript ``BN'' denotes the use of a final batch norm layer. The superscript ``N'' is the equivalent for k-nearest neighbors probing where the embedding is normalization on training split statistics.
    }
    \label{tab:results_all}
    \begin{adjustbox}{width=\textwidth}
    \begin{tabular}{lcccccccccccccccccccccccccccccc}
    \toprule
        & & \multicolumn{9}{c}{Imagenet} & \multicolumn{5}{c}{PascalVOC} & \multicolumn{5}{c}{Cifar100} & \multicolumn{5}{c}{Caltech-256} & \multicolumn{5}{c}{CUB} \\
    \cmidrule(lr){3-11} \cmidrule(lr){12-16} \cmidrule(lr){17-21} \cmidrule(lr){22-26} \cmidrule(lr){27-31}
         Method & Backbone & kNN & kNN\textsuperscript{N} & LP & LP\textsuperscript{BN} & FT & FT10\% & FT10\%\textsuperscript{BN} & FT1\% & FT1\%\textsuperscript{BN} & kNN & kNN\textsuperscript{N}  & LP & LP\textsuperscript{BN} & FT & kNN & kNN\textsuperscript{N}  & LP & LP\textsuperscript{BN} & FT & kNN & kNN\textsuperscript{N}  & LP & LP\textsuperscript{BN} & FT & kNN & kNN\textsuperscript{N}  & LP & LP\textsuperscript{BN} & FT \\
\midrule
Jigsaw \cite{noroozi_unsupervised_2016} & RN-50 & 12.0 & 12.4 & 28.6 & 32.5 & 72.2 & 50.9 & 53.8 & 17.0 & 22.2 & 39.0 & 39.1 & 44.7 & 53.7 & 64.6 & 26.7 & 27.6 & 27.4 & 36.7 & 77.5 & 26.7 & 27.3 & 29.2 & 42.4 & 74.0 & 3.4 & 4.1 & 5.1 & 7.8 & 52.6 \\
rotnet \cite{gidaris_unsupervised_2018} & RN-50 & 18.5 & 18.1 & 35.3 & 41.0 & 72.4 & 52.7 & 55.2 & 20.1 & 24.4 & 42.0 & 44.0 & 53.2 & 59.0 & 65.0 & 38.5 & 38.4 & 37.5 & 46.9 & 79.6 & 34.5 & 34.9 & 41.5 & 55.6 & 74.9 & 5.2 & 5.3 & 7.6 & 13.3 & 60.1 \\
npid \cite{wu_unsupervised_2018} & RN-50 & 36.7 & 37.1 & 31.5 & 51.0 & 71.7 & 50.5 & 56.2 & 10.2 & 29.9 & 55.4 & 54.9 & 32.5 & 67.7 & 59.8 & 40.4 & 39.9 & 23.3 & 56.5 & 78.6 & 46.5 & 46.3 & 16.0 & 67.8 & 68.5 & 15.9 & 15.7 & 10.7 & 29.0 & 32.3 \\
Sela-v2 \cite{asano_self-labelling_2020} & RN-50 & 60.4 & 61.2 & 70.4 & 70.3 & 75.9 & 68.4 & 68.6 & 50.5 & 55.5 & 73.3 & 68.7 & 82.1 & 83.4 & 87.9 & 59.2 & 57.6 & 37.9 & 46.5 & 85.6 & 68.6 & 68.0 & 74.2 & 80.7 & 87.7 & 17.1 & 19.7 & 21.1 & 41.7 & 68.4 \\
npid++ \cite{wu_unsupervised_2018, misra_self-supervised_2019} & RN-50 & 36.8 & 38.4 & 54.0 & 52.5 & 72.7 & 56.7 & 58.7 & 27.4 & 32.8 & 55.4 & 54.5 & 70.5 & 68.8 & 74.7 & 39.0 & 38.5 & 43.6 & 54.6 & 82.4 & 43.5 & 43.0 & 63.9 & 67.4 & 80.3 & 14.4 & 15.0 & 22.2 & 30.1 & 71.7 \\
PIRL \cite{misra_self-supervised_2019} & RN-50 & 49.0 & 49.5 & 58.1 & 58.7 & 73.8 & 59.9 & 60.9 & 34.3 & 39.5 & 64.5 & 61.3 & 75.1 & 73.0 & 79.1 & 47.7 & 45.3 & 35.8 & 49.2 & 83.4 & 57.4 & 58.2 & 69.9 & 72.7 & 83.2 & 16.2 & 16.4 & 21.5 & 29.3 & 71.3 \\
clusterfit \cite{yan_clusterfit_2019} & RN-50 & 35.0 & 35.7 & 48.6 & 49.8 & 72.0 & 54.3 & 56.2 & 25.2 & 30.3 & 54.0 & 50.2 & 66.2 & 66.4 & 70.1 & 49.4 & 48.9 & 62.0 & 62.2 & 81.0 & 42.6 & 48.6 & 66.6 & 66.2 & 79.0 & 7.2 & 10.3 & 18.0 & 19.3 & 58.4 \\
Deepcluster-v2 \cite{caron_deep_2018,caron_unsupervised_2020} & RN-50 & 64.8 & 65.7 & 74.7 & 73.5 & 76.5 & 69.6 & 70.0 & 50.0 & 57.4 & 75.1 & 75.4 & 85.7 & 85.4 & 87.5 & 63.1 & 63.0 & 45.6 & 48.6 & 85.9 & 73.2 & 73.7 & 83.1 & 86.1 & 89.2 & 27.1 & 33.7 & 39.0 & 56.5 & 73.0 \\
SwAV \cite{caron_unsupervised_2020} & RN-50 & 63.2 & 64.3 & 74.1 & 73.2 & 76.3 & 69.0 & 70.3 & 46.0 & 56.7 & 73.9 & 74.2 & 84.3 & 84.9 & 86.5 & 62.2 & 62.7 & 43.8 & 47.4 & 85.5 & 72.0 & 72.7 & 80.7 & 84.7 & 88.3 & 24.1 & 30.3 & 30.6 & 52.9 & 71.4 \\
SimCLR \cite{chen_simple_2020} & RN-50 & 58.1 & 57.2 & 66.9 & 68.1 & 74.6 & 65.8 & 66.6 & 47.6 & 53.6 & 73.1 & 64.2 & 81.4 & 81.9 & 87.2 & 55.6 & 51.9 & 33.5 & 50.9 & 85.1 & 64.5 & 64.0 & 70.4 & 79.9 & 86.1 & 18.5 & 19.1 & 18.9 & 37.1 & 63.2 \\
MoCo v2 \cite{chen_improved_2020} & RN-50 & 59.4 & 58.7 & 60.0 & 70.4 & 75.0 & 61.6 & 63.9 & 27.2 & 48.9 & 78.3 & 75.9 & 67.0 & 84.2 & 79.5 & 58.7 & 57.3 & 29.3 & 66.5 & 83.9 & 70.4 & 70.2 & 39.0 & 84.0 & 83.9 & 20.4 & 23.4 & 13.0 & 45.6 & 55.3 \\
SimSiam \cite{chen_exploring_2020} & RN-50 & 57.5 & 57.9 & 38.1 & 69.2 & 74.3 & 53.7 & 61.7 & 8.5 & 40.2 & 76.0 & 72.5 & 16.0 & 83.0 & 54.8 & 61.4 & 58.3 & 20.4 & 61.2 & 79.3 & 69.7 & 68.1 & 7.8 & 83.9 & 70.9 & 22.4 & 24.6 & 9.6 & 48.6 & 26.4 \\
BYOL \cite{grill_bootstrap_2020} & RN-50 & 63.2 & 64.0 & 72.7 & 72.1 & 75.6 & 68.2 & 68.4 & 50.0 & 56.3 & 75.6 & 74.3 & 84.3 & 83.8 & 86.6 & 62.7 & 59.4 & 54.2 & 60.5 & 85.8 & 71.0 & 69.5 & 81.8 & 85.3 & 87.5 & 35.3 & 39.0 & 40.2 & 58.2 & 73.4 \\
Barlow Twins \cite{zbontar_barlow_2021} & RN-50 & 61.8 & 62.5 & 71.5 & 70.8 & 74.2 & 66.2 & 65.2 & 47.4 & 51.4 & 76.1 & 73.7 & 83.8 & 83.2 & 84.6 & 61.6 & 58.4 & 57.1 & 57.7 & 84.1 & 73.0 & 71.6 & 81.9 & 83.4 & 86.0 & 30.4 & 36.2 & 42.2 & 56.0 & 61.2 \\
DenseCL \cite{wang_dense_2020} & RN-50 & 48.2 & 48.8 & 49.3 & 63.1 & 74.5 & 59.3 & 61.9 & 26.8 & 42.7 & 75.6 & 73.2 & 73.2 & 80.5 & 79.7 & 47.4 & 49.8 & 27.1 & 56.5 & 83.7 & 62.0 & 62.5 & 49.6 & 77.0 & 82.6 & 14.9 & 16.9 & 10.4 & 30.9 & 58.8 \\
DINO \cite{caron_emerging_2021} & RN-50 & 64.1 & 64.2 & 74.4 & 73.3 & 76.0 & 68.4 & 69.5 & 45.2 & 55.5 & 75.7 & 74.4 & 83.5 & 82.8 & 86.2 & 58.0 & 56.5 & 46.0 & 49.5 & 85.3 & 73.1 & 71.7 & 81.5 & 85.0 & 88.3 & 31.3 & 34.8 & 32.0 & 53.0 & 71.6 \\
MoCo v3 \cite{chen_empirical_2021} & RN-50 & 66.2 & 66.4 & 74.5 & 73.4 & 76.2 & 69.4 & 70.3 & 51.3 & 59.1 & 76.7 & 73.8 & 86.1 & 84.8 & 87.2 & 64.4 & 59.2 & 58.9 & 61.5 & 86.2 & 73.0 & 68.8 & 85.2 & 86.3 & 88.5 & 43.9 & 41.9 & 49.0 & 60.4 & 76.2 \\
DINO \cite{caron_emerging_2021} & ViT-B/16 & 74.7 & 74.9 & 76.7 & 77.6 & 81.3 & 75.4 & 76.6 & 65.6 & 69.2 & 86.5 & 86.5 & 87.6 & 88.6 & 87.7 & 76.5 & 77.0 & 82.6 & 83.0 & 90.6 & 85.7 & 86.0 & 90.6 & 91.2 & 92.4 & 58.6 & 59.6 & 78.2 & 79.2 & 83.2 \\
iBOT \cite{zhou_ibot_2021} & ViT-B/16 & 75.8 & 76.0 & 78.2 & 78.9 & 82.5 & 77.7 & 78.7 & 68.8 & 71.8 & 87.0 & 86.8 & 88.1 & 89.7 & 89.5 & 76.7 & 76.9 & 82.8 & 83.2 & 92.2 & 86.5 & 87.0 & 91.3 & 91.8 & 93.4 & 57.4 & 59.0 & 78.2 & 78.8 & 84.1 \\
MoCo v3 \cite{chen_empirical_2021} & ViT-B/16 & 70.3 & 70.9 & 73.9 & 76.3 & 80.8 & 74.4 & 74.8 & 34.6 & 66.0 & 81.8 & 82.5 & 80.9 & 88.3 & 81.9 & 75.2 & 75.9 & 76.5 & 83.3 & 88.3 & 79.6 & 80.5 & 79.5 & 89.5 & 87.7 & 43.4 & 45.3 & 39.4 & 71.8 & 59.0 \\
MAE \cite{he_masked_2021} & ViT-B/16 & 26.6 & 47.7 & 58.7 & 66.7 & 79.3 & 68.0 & 69.7 & 42.4 & 54.2 & 34.5 & 63.2 & 50.5 & 80.3 & 86.4 & 22.3 & 48.7 & 42.2 & 66.7 & 85.3 & 28.7 & 60.0 & 34.6 & 80.5 & 86.6 & 5.5 & 16.5 & 7.1 & 44.5 & 70.2 \\
MaskFeat \cite{wei_masked_2022} & ViT-B/16 & 13.7 & 30.0 & 3.7 & 51.7 & 76.4 & 56.5 & 64.8 & 7.7 & 43.7 & 34.6 & 51.8 & 16.0 & 67.0 & 55.2 & 29.8 & 44.2 & 9.6 & 58.7 & 84.2 & 21.8 & 33.7 & 6.4 & 60.1 & 76.1 & 4.0 & 5.6 & 1.5 & 15.6 & 29.5 \\
BEiT v2 \cite{peng_beit_2022} & ViT-B/16 & 69.1 & 70.0 & 78.4 & 78.2 & 84.8 & 80.5 & 80.7 & 68.2 & 72.8 & 82.1 & 82.3 & 89.1 & 88.7 & 91.8 & 74.0 & 74.4 & 83.0 & 83.0 & 92.3 & 79.0 & 79.7 & 90.2 & 90.5 & 94.7 & 19.5 & 20.7 & 50.7 & 52.9 & 84.7 \\
MILAN \cite{hou_milan_2022} & ViT-B/16 & 72.9 & 73.2 & 76.4 & 78.8 & 84.4 & 78.9 & 79.0 & 67.3 & 69.4 & 87.3 & 87.0 & 89.0 & 90.0 & 91.9 & 70.6 & 71.2 & 70.9 & 79.7 & 90.0 & 87.7 & 87.8 & 88.2 & 92.4 & 94.5 & 56.8 & 57.7 & 52.0 & 75.4 & 83.1 \\
EVA \cite{fang_eva_2022} & ViT-B/16 & 47.1 & 50.9 & 60.5 & 68.7 & 82.1 & 71.6 & 72.9 & 41.1 & 52.5 & 61.0 & 65.5 & 47.5 & 79.3 & 83.4 & 55.2 & 60.1 & 54.6 & 76.6 & 87.4 & 53.9 & 59.5 & 29.8 & 82.8 & 89.0 & 11.6 & 13.8 & 9.0 & 39.1 & 65.3 \\
PixMIM \cite{liu_pixmim_2023} & ViT-B/16 & 39.4 & 49.5 & 59.3 & 66.5 & 79.2 & 63.3 & 66.1 & 25.9 & 44.3 & 51.9 & 65.3 & 50.4 & 81.5 & 80.0 & 43.4 & 57.5 & 50.0 & 71.8 & 81.1 & 47.6 & 62.3 & 35.8 & 81.1 & 81.5 & 9.6 & 15.5 & 8.0 & 40.8 & 48.2 \\
    \end{tabular}
\end{adjustbox}
    \begin{adjustbox}{width=\textwidth}
    \begin{tabular}{lccccccccccccccccccccccccccccccc}
    \toprule
    & & \multicolumn{30}{c}{Imagenet-D} \\
        & & \multicolumn{5}{c}{Clipart} & \multicolumn{5}{c}{Infograph} & \multicolumn{5}{c}{Painting} & \multicolumn{5}{c}{Quickdraw} & \multicolumn{5}{c}{Real} & \multicolumn{5}{c}{Sketch} \\
    \cmidrule(lr){3-7} \cmidrule(lr){8-12} \cmidrule(lr){13-17} \cmidrule(lr){18-22} \cmidrule(lr){23-27} \cmidrule(lr){28-32}
         Method & Backbone & kNN & kNN\textsuperscript{N} & LP & LP\textsuperscript{BN} & FT & kNN & kNN\textsuperscript{N}  & LP & LP\textsuperscript{BN} & FT & kNN & kNN\textsuperscript{N} & LP & LP\textsuperscript{BN} & FT & kNN & kNN\textsuperscript{N} & LP & LP\textsuperscript{BN} & FT & kNN  & kNN\textsuperscript{N} & LP & LP\textsuperscript{BN} & FT & kNN & kNN\textsuperscript{N} & LP & LP\textsuperscript{BN} & FT \\
\midrule
Jigsaw \cite{noroozi_unsupervised_2016} & RN-50 & 16.0 & 16.7 & 22.1 & 35.7 & 65.0 & 8.8 & 8.7 & 9.8 & 13.4 & 31.3 & 15.0 & 15.1 & 23.1 & 32.1 & 57.5 & 30.9 & 32.5 & 16.4 & 26.5 & 71.7 & 29.2 & 29.4 & 43.0 & 53.0 & 76.2 & 14.5 & 15.1 & 17.7 & 27.1 & 59.1 \\
rotnet \cite{gidaris_unsupervised_2018} & RN-50 & 23.7 & 24.7 & 34.5 & 46.9 & 68.2 & 9.0 & 10.0 & 12.9 & 17.8 & 34.6 & 21.7 & 21.7 & 34.0 & 41.9 & 60.8 & 35.4 & 35.0 & 35.2 & 49.7 & 72.0 & 36.3 & 36.5 & 52.2 & 60.3 & 77.4 & 20.5 & 21.4 & 29.3 & 40.4 & 62.2 \\
npid \cite{wu_unsupervised_2018} & RN-50 & 29.1 & 28.4 & 10.6 & 47.3 & 65.3 & 14.0 & 14.2 & 5.3 & 19.1 & 31.4 & 30.7 & 31.0 & 16.8 & 45.2 & 57.4 & 40.6 & 37.4 & 8.4 & 45.1 & 70.7 & 50.5 & 50.8 & 36.6 & 64.3 & 76.1 & 25.9 & 25.8 & 9.6 & 38.5 & 59.3 \\
Sela-v2 \cite{asano_self-labelling_2020} & RN-50 & 35.4 & 34.3 & 48.8 & 55.2 & 74.1 & 17.8 & 18.9 & 22.8 & 25.7 & 42.4 & 45.3 & 45.9 & 56.6 & 59.1 & 70.2 & 32.1 & 27.2 & 33.2 & 42.8 & 70.5 & 66.9 & 67.5 & 75.3 & 75.9 & 82.3 & 32.1 & 32.4 & 44.0 & 48.7 & 67.7 \\
npid++ \cite{wu_unsupervised_2018, misra_self-supervised_2019} & RN-50 & 23.5 & 22.1 & 40.8 & 45.5 & 69.3 & 13.8 & 13.8 & 17.2 & 18.5 & 35.2 & 29.2 & 29.2 & 43.6 & 43.5 & 62.0 & 26.4 & 19.7 & 29.4 & 43.3 & 71.6 & 49.7 & 49.9 & 65.8 & 64.6 & 78.4 & 18.9 & 18.7 & 29.1 & 34.0 & 63.3 \\
PIRL \cite{misra_self-supervised_2019} & RN-50 & 34.9 & 34.0 & 47.8 & 55.4 & 71.4 & 16.4 & 17.7 & 20.5 & 23.9 & 37.5 & 39.7 & 39.8 & 50.1 & 52.6 & 64.7 & 33.0 & 24.6 & 32.2 & 39.8 & 71.5 & 59.5 & 59.5 & 69.2 & 70.9 & 79.6 & 32.0 & 31.2 & 40.9 & 46.7 & 64.6 \\
clusterfit \cite{yan_clusterfit_2019} & RN-50 & 34.6 & 35.4 & 52.0 & 52.5 & 68.7 & 14.5 & 15.3 & 20.0 & 20.4 & 33.6 & 35.0 & 36.1 & 46.4 & 47.1 & 61.1 & 41.1 & 41.2 & 51.1 & 51.4 & 69.9 & 52.7 & 53.0 & 65.6 & 66.5 & 78.2 & 30.0 & 30.8 & 43.2 & 43.7 & 61.6 \\
Deepcluster-v2 \cite{caron_deep_2018,caron_unsupervised_2020} & RN-50 & 40.0 & 40.6 & 59.8 & 60.3 & 74.7 & 21.1 & 22.3 & 28.7 & 28.9 & 42.1 & 49.2 & 50.3 & 64.6 & 63.5 & 70.7 & 33.7 & 34.4 & 44.1 & 47.4 & 71.0 & 70.8 & 71.3 & 80.3 & 78.7 & 82.8 & 35.9 & 36.5 & 52.3 & 53.3 & 68.4 \\
SwAV \cite{caron_unsupervised_2020} & RN-50 & 38.3 & 39.1 & 57.5 & 59.6 & 74.4 & 20.1 & 21.8 & 27.3 & 28.4 & 42.0 & 48.1 & 49.7 & 63.0 & 62.9 & 70.4 & 32.6 & 31.4 & 45.5 & 50.5 & 71.3 & 70.2 & 70.8 & 79.3 & 78.1 & 82.5 & 34.7 & 35.8 & 50.8 & 53.3 & 68.4 \\
SimCLR \cite{chen_simple_2020} & RN-50 & 33.4 & 34.6 & 38.6 & 55.9 & 73.8 & 17.2 & 18.6 & 16.2 & 24.9 & 40.1 & 44.1 & 44.3 & 51.2 & 57.9 & 68.6 & 27.6 & 18.0 & 26.1 & 46.4 & 71.1 & 64.7 & 64.0 & 70.1 & 75.0 & 81.3 & 32.1 & 32.2 & 36.3 & 49.9 & 66.7 \\
MoCo v2 \cite{chen_improved_2020} & RN-50 & 36.6 & 38.9 & 18.7 & 62.5 & 71.5 & 18.1 & 20.3 & 8.1 & 29.7 & 37.2 & 49.0 & 49.2 & 33.4 & 62.0 & 65.1 & 33.1 & 30.8 & 13.2 & 47.8 & 72.2 & 66.5 & 66.8 & 59.3 & 78.0 & 80.3 & 35.9 & 36.9 & 17.6 & 55.9 & 65.2 \\
SimSiam \cite{chen_exploring_2020} & RN-50 & 39.2 & 41.3 & 1.0 & 63.9 & 67.3 & 18.8 & 20.8 & 1.6 & 29.6 & 32.7 & 48.8 & 49.0 & 1.5 & 63.0 & 60.5 & 38.8 & 35.4 & 11.6 & 54.9 & 71.4 & 66.9 & 67.4 & 18.4 & 77.1 & 78.0 & 38.2 & 38.9 & 2.5 & 57.2 & 61.0 \\
BYOL \cite{grill_bootstrap_2020} & RN-50 & 40.2 & 42.2 & 58.2 & 63.5 & 75.2 & 19.1 & 21.6 & 25.3 & 28.9 & 42.2 & 49.8 & 49.8 & 62.5 & 63.7 & 70.4 & 36.9 & 31.1 & 38.3 & 47.5 & 71.3 & 69.2 & 69.5 & 78.4 & 78.3 & 82.4 & 39.4 & 39.8 & 52.6 & 56.7 & 68.9 \\
Barlow Twins \cite{zbontar_barlow_2021} & RN-50 & 40.4 & 41.1 & 62.9 & 63.0 & 71.3 & 20.1 & 21.1 & 28.4 & 28.1 & 36.7 & 49.9 & 49.6 & 63.7 & 62.1 & 67.6 & 38.5 & 35.0 & 44.9 & 47.1 & 68.7 & 69.5 & 70.0 & 78.8 & 77.2 & 81.1 & 39.8 & 39.5 & 55.6 & 55.5 & 64.7 \\
DenseCL \cite{wang_dense_2020} & RN-50 & 31.1 & 31.9 & 21.0 & 54.5 & 71.4 & 15.2 & 15.8 & 10.3 & 24.1 & 36.5 & 41.8 & 42.6 & 33.8 & 55.4 & 65.1 & 29.4 & 31.9 & 13.6 & 48.5 & 71.9 & 59.5 & 60.2 & 54.5 & 73.1 & 80.0 & 30.0 & 30.5 & 20.0 & 47.6 & 64.9 \\
DINO \cite{caron_emerging_2021} & RN-50 & 37.9 & 38.8 & 60.3 & 61.2 & 73.6 & 20.3 & 22.2 & 28.6 & 28.5 & 41.8 & 48.9 & 49.1 & 64.3 & 63.3 & 70.1 & 33.9 & 32.3 & 43.9 & 48.1 & 70.9 & 71.0 & 70.7 & 80.2 & 78.8 & 82.5 & 36.0 & 36.3 & 52.5 & 53.3 & 67.7 \\
MoCo v3 \cite{chen_empirical_2021} & RN-50 & 46.8 & 45.6 & 64.9 & 65.4 & 76.0 & 21.2 & 22.3 & 29.4 & 30.4 & 43.1 & 53.2 & 51.7 & 65.7 & 64.9 & 71.5 & 40.4 & 35.6 & 47.5 & 50.9 & 71.6 & 70.7 & 70.2 & 80.1 & 78.7 & 83.0 & 45.1 & 43.5 & 58.6 & 59.1 & 69.9 \\
DINO \cite{caron_emerging_2021} & ViT-B/16 & 53.4 & 53.9 & 67.7 & 68.1 & 77.8 & 29.6 & 29.8 & 37.2 & 37.9 & 45.4 & 63.1 & 63.3 & 67.3 & 68.5 & 73.7 & 39.5 & 40.3 & 56.7 & 57.3 & 71.5 & 78.8 & 79.0 & 81.1 & 81.9 & 86.0 & 48.4 & 48.7 & 58.7 & 59.1 & 71.0 \\
iBOT \cite{zhou_ibot_2021} & ViT-B/16 & 55.2 & 55.1 & 69.5 & 70.0 & 80.2 & 30.6 & 30.2 & 38.3 & 39.1 & 47.8 & 63.6 & 63.7 & 68.5 & 69.7 & 75.9 & 41.6 & 42.5 & 60.1 & 60.4 & 72.7 & 79.0 & 79.2 & 81.3 & 82.4 & 87.0 & 48.2 & 48.2 & 59.5 & 60.1 & 73.9 \\
MoCo v3 \cite{chen_empirical_2021} & ViT-B/16 & 50.5 & 51.0 & 39.2 & 68.1 & 74.0 & 26.1 & 26.2 & 19.6 & 35.2 & 40.8 & 57.7 & 58.4 & 53.9 & 66.4 & 72.4 & 42.1 & 42.8 & 32.4 & 59.3 & 68.6 & 75.3 & 75.7 & 75.4 & 81.1 & 85.3 & 43.6 & 44.1 & 35.2 & 56.8 & 67.6 \\
MAE \cite{he_masked_2021} & ViT-B/16 & 21.1 & 41.4 & 25.9 & 61.7 & 73.4 & 10.4 & 18.8 & 11.2 & 28.4 & 40.4 & 15.6 & 39.9 & 32.0 & 57.7 & 69.5 & 40.9 & 44.7 & 28.8 & 57.4 & 70.7 & 35.2 & 61.0 & 60.4 & 76.2 & 83.2 & 22.9 & 34.2 & 20.3 & 50.4 & 66.1 \\
MaskFeat \cite{wei_masked_2022} & ViT-B/16 & 14.3 & 19.8 & 2.2 & 43.0 & 63.1 & 7.5 & 10.7 & 2.0 & 16.5 & 26.7 & 12.0 & 24.2 & 3.3 & 43.8 & 58.9 & 22.6 & 31.3 & 2.0 & 42.5 & 67.2 & 25.3 & 40.8 & 4.2 & 61.4 & 78.2 & 12.9 & 18.5 & 2.5 & 32.8 & 55.7 \\
BEiT v2 \cite{peng_beit_2022} & ViT-B/16 & 57.8 & 57.9 & 72.8 & 72.1 & 83.3 & 26.1 & 25.8 & 39.2 & 38.9 & 51.7 & 59.8 & 60.0 & 70.5 & 69.8 & 79.6 & 43.7 & 44.7 & 56.4 & 58.7 & 73.5 & 78.2 & 78.4 & 83.7 & 83.5 & 88.8 & 48.4 & 48.7 & 63.1 & 62.9 & 76.3 \\
MILAN \cite{hou_milan_2022} & ViT-B/16 & 67.5 & 68.1 & 66.5 & 75.2 & 82.8 & 36.4 & 36.7 & 34.7 & 44.8 & 53.4 & 65.7 & 66.0 & 67.5 & 72.8 & 79.8 & 48.5 & 49.0 & 40.0 & 59.3 & 71.5 & 82.9 & 83.0 & 84.2 & 86.0 & 88.8 & 57.0 & 57.5 & 55.6 & 66.0 & 75.4 \\
EVA \cite{fang_eva_2022} & ViT-B/16 & 36.9 & 40.9 & 28.8 & 62.9 & 74.6 & 18.8 & 20.3 & 12.8 & 31.0 & 43.0 & 37.4 & 41.5 & 35.4 & 59.3 & 72.5 & 49.4 & 50.8 & 34.4 & 58.4 & 70.9 & 61.0 & 64.8 & 64.7 & 77.9 & 85.6 & 30.0 & 32.4 & 21.9 & 51.1 & 67.0 \\
PixMIM \cite{liu_pixmim_2023} & ViT-B/16 & 29.6 & 42.1 & 26.9 & 62.1 & 68.0 & 14.9 & 19.5 & 11.9 & 28.7 & 36.6 & 28.8 & 42.8 & 34.5 & 58.2 & 66.2 & 42.1 & 44.5 & 28.0 & 55.3 & 69.1 & 51.6 & 63.0 & 61.9 & 76.1 & 82.1 & 25.8 & 34.2 & 20.4 & 50.4 & 61.2 \\
    \bottomrule
    \end{tabular}
\end{adjustbox}
    \begin{adjustbox}{width=.6\textwidth}
    \begin{tabular}{lcccccccccccccccc}
    & & \multicolumn{15}{c}{iNaturalist mini} \\
        & & \multicolumn{5}{c}{Target: Family} & \multicolumn{5}{c}{Target: Genus} & \multicolumn{5}{c}{Target: Species} \\
    \cmidrule(lr){3-7} \cmidrule(lr){8-12} \cmidrule(lr){13-17}
         Method & Backbone & kNN & kNN\textsuperscript{N} & LP & LP\textsuperscript{BN} & FT & kNN & kNN\textsuperscript{N}  & LP & LP\textsuperscript{BN} & FT & kNN & kNN\textsuperscript{N}  & LP & LP\textsuperscript{BN} & FT \\
\midrule
Jigsaw \cite{noroozi_unsupervised_2016} & RN-50 & 11.4 & 11.6 & 11.9 & 19.6 & 72.7 & 2.2 & 2.2 & 4.1 & 6.3 & 42.7 & 1.4 & 1.6 & 2.8 & 3.9 & 34.7 \\
rotnet \cite{gidaris_unsupervised_2018} & RN-50 & 13.4 & 13.4 & 19.9 & 23.1 & 72.9 & 3.0 & 3.0 & 6.0 & 9.7 & 46.3 & 2.1 & 2.3 & 4.2 & 6.4 & 38.0 \\
npid \cite{wu_unsupervised_2018} & RN-50 & 21.0 & 21.0 & 17.6 & 31.3 & 71.5 & 8.0 & 8.2 & 5.5 & 17.2 & 42.1 & 5.5 & 5.8 & 3.6 & 12.1 & 36.5 \\
Sela-v2 \cite{asano_self-labelling_2020} & RN-50 & 31.7 & 33.2 & 39.6 & 42.7 & 76.5 & 14.1 & 16.4 & 21.1 & 29.9 & 60.5 & 9.7 & 10.7 & 14.3 & 20.5 & 48.6 \\
npid++ \cite{wu_unsupervised_2018, misra_self-supervised_2019} & RN-50 & 20.8 & 21.2 & 31.0 & 31.3 & 75.5 & 7.8 & 8.1 & 15.6 & 17.4 & 54.1 & 5.5 & 5.6 & 11.2 & 12.2 & 45.2 \\
PIRL \cite{misra_self-supervised_2019} & RN-50 & 26.9 & 27.1 & 19.4 & 35.3 & 75.7 & 11.0 & 11.6 & 18.2 & 21.3 & 54.7 & 7.6 & 7.9 & 12.9 & 15.1 & 44.8 \\
clusterfit \cite{yan_clusterfit_2019} & RN-50 & 19.9 & 20.5 & 29.2 & 29.3 & 72.8 & 6.5 & 7.0 & 12.8 & 13.2 & 44.7 & 4.2 & 4.6 & 8.5 & 8.8 & 34.5 \\
Deepcluster-v2 \cite{caron_deep_2018,caron_unsupervised_2020} & RN-50 & 38.1 & 40.3 & 49.8 & 50.5 & 77.7 & 20.1 & 22.9 & 36.2 & 39.6 & 62.2 & 13.7 & 16.2 & 26.3 & 29.2 & 50.8 \\
SwAV \cite{caron_unsupervised_2020} & RN-50 & 36.0 & 38.6 & 47.0 & 48.4 & 77.7 & 17.1 & 21.2 & 30.2 & 36.2 & 61.6 & 12.2 & 14.8 & 21.3 & 26.1 & 49.9 \\
SimCLR \cite{chen_simple_2020} & RN-50 & 29.1 & 29.4 & 34.0 & 40.4 & 75.2 & 13.4 & 13.9 & 15.9 & 26.2 & 55.0 & 8.8 & 9.7 & 10.5 & 17.5 & 44.0 \\
MoCo v2 \cite{chen_improved_2020} & RN-50 & 32.6 & 33.6 & 28.4 & 48.1 & 77.3 & 16.5 & 17.3 & 9.3 & 34.7 & 55.1 & 10.8 & 11.8 & 6.2 & 25.3 & 46.5 \\
SimSiam \cite{chen_exploring_2020} & RN-50 & 32.8 & 33.6 & 9.5 & 45.9 & 75.6 & 15.9 & 17.7 & 8.4 & 33.0 & 45.5 & 10.7 & 12.3 & 5.0 & 24.2 & 40.3 \\
BYOL \cite{grill_bootstrap_2020} & RN-50 & 37.6 & 39.3 & 45.1 & 48.3 & 76.3 & 20.3 & 24.6 & 28.5 & 38.3 & 60.9 & 14.1 & 17.1 & 20.2 & 27.8 & 50.1 \\
Barlow Twins \cite{zbontar_barlow_2021} & RN-50 & 37.6 & 39.0 & 47.3 & 47.8 & 66.7 & 20.2 & 22.8 & 33.9 & 36.3 & 50.4 & 13.7 & 16.0 & 24.6 & 26.7 & 38.8 \\
DenseCL \cite{wang_dense_2020} & RN-50 & 26.0 & 26.7 & 25.0 & 38.5 & 76.7 & 11.3 & 12.2 & 7.6 & 24.1 & 52.5 & 7.1 & 7.7 & 5.2 & 17.1 & 44.2 \\
DINO \cite{caron_emerging_2021} & RN-50 & 40.1 & 40.5 & 50.1 & 51.3 & 77.8 & 21.8 & 23.0 & 33.4 & 38.4 & 60.7 & 14.8 & 15.6 & 23.1 & 27.3 & 49.0 \\
MoCo v3 \cite{chen_empirical_2021} & RN-50 & 40.5 & 40.0 & 47.4 & 48.6 & 77.5 & 25.8 & 26.9 & 35.2 & 40.3 & 64.4 & 18.1 & 19.4 & 26.1 & 29.7 & 53.1 \\
DINO \cite{caron_emerging_2021} & ViT-B/16 & 55.8 & 56.4 & 60.6 & 61.1 & 82.2 & 44.5 & 45.3 & 60.5 & 61.5 & 71.3 & 32.6 & 33.0 & 49.0 & 48.9 & 59.1 \\
iBOT \cite{zhou_ibot_2021} & ViT-B/16 & 53.4 & 54.1 & 60.6 & 61.2 & 84.1 & 42.0 & 43.3 & 60.6 & 61.1 & 73.9 & 31.5 & 32.1 & 47.9 & 48.5 & 61.7 \\
MoCo v3 \cite{chen_empirical_2021} & ViT-B/16 & 45.0 & 45.7 & 43.4 & 56.2 & 75.2 & 29.0 & 30.0 & 23.6 & 51.8 & 66.7 & 20.6 & 21.4 & 16.8 & 40.0 & 55.6 \\
MAE \cite{he_masked_2021} & ViT-B/16 & 13.7 & 25.7 & 30.8 & 44.5 & 79.6 & 3.0 & 10.2 & 7.8 & 31.9 & 65.3 & 2.4 & 7.2 & 5.0 & 23.5 & 54.5 \\
MaskFeat \cite{wei_masked_2022} & ViT-B/16 & 10.0 & 17.9 & 8.2 & 31.4 & 75.2 & 2.0 & 5.1 & 0.5 & 15.1 & 46.4 & 1.5 & 3.6 & 0.5 & 10.3 & 39.0 \\
BEiT v2 \cite{peng_beit_2022} & ViT-B/16 & 43.1 & 44.1 & 58.9 & 59.2 & 85.4 & 18.7 & 19.6 & 44.5 & 44.6 & 79.2 & 12.6 & 13.1 & 33.7 & 33.5 & 68.4 \\
MILAN \cite{hou_milan_2022} & ViT-B/16 & 53.5 & 54.2 & 50.7 & 61.3 & 84.5 & 36.4 & 37.2 & 34.7 & 56.2 & 74.7 & 27.0 & 27.6 & 26.4 & 45.7 & 64.4 \\
EVA \cite{fang_eva_2022} & ViT-B/16 & 27.6 & 30.8 & 34.8 & 49.4 & 82.0 & 8.8 & 11.2 & 8.6 & 30.9 & 69.9 & 6.3 & 8.2 & 5.8 & 22.0 & 59.3 \\
PixMIM \cite{liu_pixmim_2023} & ViT-B/16 & 20.5 & 27.1 & 31.4 & 44.7 & 76.9 & 6.0 & 10.7 & 8.5 & 32.1 & 61.3 & 4.0 & 7.6 & 5.6 & 22.8 & 50.9 \\
\bottomrule
\end{tabular}
\end{adjustbox}
    \end{table*}

\begin{table*}[ht!]
    \centering
    \caption{
    {\em Top-5} classification accuracies (in \%). LP: Linear Probing, FT: End-to-end Fine-tuning, FT 10\%/1\%: Few-Shot Fine-tuning on 10\%/1\% of the training data, kNN: k-nearest neighbors probing.
    The superscript ``BN'' denotes the use of a final batch norm layer. The superscript ``N'' is the equivalent for k-nearest neighbors probing where the embedding is normalization on training split statistics.
    }
    \label{tab:results_all_top5}
    \begin{adjustbox}{width=\textwidth}
    \begin{tabular}{lcccccccccccccccccccccccccccccc}
    \toprule
        & & \multicolumn{9}{c}{Imagenet} & \multicolumn{5}{c}{PascalVOC} & \multicolumn{5}{c}{Cifar100} & \multicolumn{5}{c}{Caltech-256} & \multicolumn{5}{c}{CUB} \\
    \cmidrule(lr){3-11} \cmidrule(lr){12-16} \cmidrule(lr){17-21} \cmidrule(lr){22-26} \cmidrule(lr){27-31}
         Method & Backbone & kNN & kNN\textsuperscript{N} & LP & LP\textsuperscript{BN} & FT & FT10\% & FT10\%\textsuperscript{BN} & FT1\% & FT1\%\textsuperscript{BN} & kNN & kNN\textsuperscript{N}  & LP & LP\textsuperscript{BN} & FT & kNN & kNN\textsuperscript{N}  & LP & LP\textsuperscript{BN} & FT & kNN & kNN\textsuperscript{N}  & LP & LP\textsuperscript{BN} & FT & kNN & kNN\textsuperscript{N}  & LP & LP\textsuperscript{BN} & FT \\
\midrule
Jigsaw \cite{noroozi_unsupervised_2016} & RN-50 & 24.8 & 25.4 & 52.2 & 56.4 & 90.5 & 76.6 & 77.9 & 39.3 & 45.8 & 75.6 & 75.3 & 85.1 & 86.6 & 92.1 & 51.8 & 52.7 & 54.2 & 65.6 & 94.8 & 45.3 & 46.2 & 50.1 & 64.4 & 89.7 & 11.4 & 12.8 & 16.6 & 23.1 & 82.3 \\
rotnet \cite{gidaris_unsupervised_2018} & RN-50 & 34.8 & 34.3 & 59.8 & 65.4 & 90.7 & 78.0 & 78.8 & 42.8 & 48.3 & 79.4 & 80.3 & 87.9 & 88.8 & 92.4 & 66.7 & 66.1 & 65.7 & 74.3 & 95.4 & 55.8 & 55.9 & 66.0 & 76.7 & 89.6 & 16.1 & 16.1 & 23.5 & 33.1 & 86.5 \\
npid \cite{wu_unsupervised_2018} & RN-50 & 57.9 & 58.4 & 56.5 & 74.8 & 90.2 & 76.5 & 79.7 & 27.1 & 54.8 & 87.0 & 86.5 & 70.8 & 92.3 & 86.5 & 68.2 & 66.9 & 50.2 & 82.4 & 95.2 & 68.5 & 69.1 & 33.0 & 85.2 & 85.4 & 34.9 & 35.2 & 31.4 & 56.2 & 67.2 \\
Sela-v2 \cite{asano_self-labelling_2020} & RN-50 & 80.1 & 80.4 & 90.2 & 89.7 & 93.3 & 89.4 & 89.1 & 77.1 & 80.2 & 94.8 & 92.8 & 97.7 & 98.3 & 98.9 & 84.4 & 82.5 & 67.3 & 74.7 & 98.0 & 85.7 & 86.1 & 89.2 & 92.9 & 96.8 & 38.5 & 42.2 & 47.6 & 71.3 & 91.7 \\
npid++ \cite{wu_unsupervised_2018, misra_self-supervised_2019} & RN-50 & 58.4 & 60.0 & 78.6 & 77.2 & 91.1 & 81.0 & 81.6 & 53.5 & 58.9 & 86.1 & 85.8 & 95.5 & 94.3 & 96.2 & 68.4 & 67.1 & 74.1 & 82.9 & 96.4 & 67.1 & 67.1 & 84.2 & 85.9 & 92.1 & 33.6 & 35.1 & 49.2 & 59.8 & 91.6 \\
PIRL \cite{misra_self-supervised_2019} & RN-50 & 71.4 & 71.8 & 81.9 & 81.8 & 91.8 & 83.6 & 83.6 & 61.9 & 66.4 & 91.8 & 90.1 & 95.5 & 94.5 & 97.2 & 75.4 & 71.8 & 65.3 & 77.6 & 96.5 & 78.7 & 79.4 & 87.7 & 88.8 & 93.9 & 38.5 & 37.4 & 48.6 & 58.6 & 91.5 \\
clusterfit \cite{yan_clusterfit_2019} & RN-50 & 56.2 & 57.0 & 72.8 & 74.0 & 90.5 & 78.7 & 79.7 & 48.5 & 55.4 & 86.5 & 83.0 & 92.4 & 92.2 & 93.6 & 77.2 & 77.0 & 87.2 & 87.4 & 96.0 & 65.5 & 71.4 & 85.1 & 85.3 & 91.8 & 19.9 & 24.7 & 42.5 & 44.7 & 84.9 \\
Deepcluster-v2 \cite{caron_deep_2018,caron_unsupervised_2020} & RN-50 & 83.8 & 84.2 & 92.0 & 91.2 & 93.4 & 89.8 & 89.9 & 76.3 & 81.3 & 96.2 & 96.1 & 98.5 & 98.2 & 98.6 & 86.2 & 86.3 & 74.0 & 76.7 & 98.2 & 89.2 & 89.4 & 94.5 & 95.6 & 97.2 & 53.2 & 61.5 & 68.1 & 82.7 & 93.2 \\
SwAV \cite{caron_unsupervised_2020} & RN-50 & 82.9 & 83.5 & 91.8 & 91.2 & 93.4 & 89.6 & 90.0 & 73.4 & 81.1 & 96.1 & 96.1 & 98.5 & 98.6 & 98.8 & 86.2 & 86.3 & 72.5 & 75.6 & 97.9 & 88.6 & 89.2 & 93.4 & 95.2 & 96.9 & 48.5 & 57.5 & 59.1 & 80.3 & 92.2 \\
SimCLR \cite{chen_simple_2020} & RN-50 & 78.1 & 76.7 & 87.9 & 88.5 & 92.5 & 87.9 & 88.0 & 75.4 & 79.6 & 94.1 & 91.5 & 97.6 & 98.3 & 99.3 & 82.2 & 78.3 & 61.9 & 79.3 & 98.0 & 83.8 & 83.0 & 88.1 & 92.8 & 96.1 & 41.7 & 39.7 & 45.1 & 67.2 & 90.4 \\
MoCo v2 \cite{chen_improved_2020} & RN-50 & 80.6 & 79.3 & 83.8 & 89.8 & 92.3 & 84.8 & 85.5 & 53.0 & 74.8 & 96.0 & 95.5 & 91.9 & 98.6 & 96.7 & 83.9 & 82.5 & 60.1 & 90.1 & 96.8 & 86.8 & 86.8 & 65.6 & 94.6 & 93.3 & 45.3 & 48.4 & 39.3 & 74.8 & 87.5 \\
SimSiam \cite{chen_exploring_2020} & RN-50 & 79.0 & 79.0 & 67.0 & 88.9 & 91.8 & 79.3 & 84.0 & 24.2 & 66.5 & 95.5 & 94.8 & 47.9 & 97.8 & 85.4 & 85.9 & 83.5 & 46.5 & 86.9 & 95.1 & 86.9 & 86.5 & 10.3 & 94.6 & 87.0 & 46.9 & 49.2 & 26.0 & 77.3 & 65.4 \\
BYOL \cite{grill_bootstrap_2020} & RN-50 & 82.1 & 82.3 & 90.9 & 90.5 & 93.0 & 89.1 & 89.0 & 76.4 & 80.7 & 95.8 & 95.1 & 98.5 & 98.1 & 99.0 & 87.0 & 84.3 & 82.1 & 86.0 & 98.1 & 87.8 & 87.3 & 93.6 & 95.4 & 96.5 & 62.7 & 67.2 & 71.0 & 84.3 & 93.5 \\
Barlow Twins \cite{zbontar_barlow_2021} & RN-50 & 81.1 & 81.2 & 89.9 & 89.4 & 91.9 & 87.4 & 86.5 & 73.9 & 76.3 & 95.9 & 95.2 & 98.0 & 97.6 & 98.3 & 85.5 & 82.9 & 83.5 & 83.4 & 97.7 & 88.4 & 88.1 & 93.8 & 94.6 & 95.7 & 58.4 & 64.0 & 71.4 & 82.2 & 86.8 \\
DenseCL \cite{wang_dense_2020} & RN-50 & 72.5 & 72.8 & 76.6 & 85.6 & 92.0 & 83.7 & 84.6 & 54.6 & 69.8 & 95.7 & 96.2 & 96.1 & 97.2 & 97.6 & 76.4 & 77.6 & 56.9 & 84.1 & 96.8 & 83.4 & 83.5 & 76.8 & 91.2 & 93.3 & 35.4 & 38.1 & 31.3 & 61.3 & 88.6 \\
DINO \cite{caron_emerging_2021} & RN-50 & 83.8 & 83.4 & 92.1 & 91.3 & 93.3 & 89.4 & 89.8 & 72.9 & 80.6 & 96.2 & 96.0 & 98.5 & 98.2 & 99.0 & 82.8 & 81.4 & 74.8 & 78.3 & 98.0 & 89.5 & 88.8 & 93.7 & 95.2 & 97.0 & 57.1 & 62.0 & 60.7 & 80.4 & 92.6 \\
MoCo v3 \cite{chen_empirical_2021} & RN-50 & 84.2 & 83.8 & 91.8 & 91.0 & 93.4 & 89.7 & 90.0 & 77.7 & 82.5 & 96.5 & 95.0 & 98.4 & 97.9 & 98.7 & 87.5 & 82.9 & 85.4 & 86.9 & 98.2 & 89.5 & 88.0 & 95.3 & 95.5 & 96.8 & 73.0 & 70.4 & 78.4 & 86.4 & 94.4 \\
DINO \cite{caron_emerging_2021} & ViT-B/16 & 91.1 & 91.2 & 93.1 & 93.6 & 95.6 & 92.8 & 93.7 & 87.9 & 90.2 & 97.5 & 97.5 & 99.2 & 99.1 & 99.0 & 94.1 & 93.7 & 97.2 & 97.3 & 99.0 & 96.1 & 96.3 & 97.5 & 97.9 & 98.1 & 85.7 & 86.0 & 95.0 & 95.1 & 96.3 \\
iBOT \cite{zhou_ibot_2021} & ViT-B/16 & 91.7 & 91.7 & 93.8 & 94.2 & 96.2 & 94.0 & 94.5 & 89.9 & 91.8 & 97.9 & 98.2 & 99.2 & 99.1 & 99.0 & 93.7 & 93.6 & 97.2 & 97.2 & 99.3 & 96.1 & 96.4 & 97.8 & 98.0 & 98.4 & 84.2 & 85.0 & 94.9 & 95.3 & 96.4 \\
MoCo v3 \cite{chen_empirical_2021} & ViT-B/16 & 88.3 & 88.7 & 91.9 & 93.0 & 95.5 & 92.5 & 92.5 & 60.2 & 88.0 & 96.8 & 96.5 & 97.5 & 99.3 & 98.2 & 93.5 & 93.7 & 95.1 & 97.1 & 98.9 & 92.8 & 93.3 & 93.5 & 97.2 & 96.3 & 72.6 & 74.0 & 72.8 & 92.1 & 86.9 \\
MAE \cite{he_masked_2021} & ViT-B/16 & 44.1 & 68.6 & 81.0 & 86.8 & 94.6 & 88.8 & 89.6 & 71.3 & 79.7 & 71.1 & 89.8 & 82.1 & 97.5 & 98.6 & 46.4 & 75.4 & 71.7 & 90.4 & 98.2 & 47.1 & 79.0 & 59.1 & 92.6 & 95.9 & 16.4 & 37.3 & 20.1 & 73.9 & 92.6 \\
MaskFeat \cite{wei_masked_2022} & ViT-B/16 & 26.0 & 48.9 & 10.7 & 75.5 & 92.6 & 80.5 & 86.0 & 23.7 & 71.0 & 72.5 & 83.4 & 52.0 & 92.7 & 88.0 & 56.2 & 71.5 & 28.6 & 85.9 & 96.7 & 38.0 & 51.9 & 11.5 & 79.7 & 89.5 & 12.7 & 17.0 & 6.4 & 36.6 & 66.5 \\
BEiT v2 \cite{peng_beit_2022} & ViT-B/16 & 86.9 & 87.2 & 94.3 & 94.1 & 97.4 & 95.8 & 95.9 & 90.5 & 92.4 & 97.6 & 97.6 & 99.3 & 99.2 & 99.8 & 92.3 & 92.2 & 97.2 & 97.2 & 99.3 & 92.6 & 92.2 & 97.5 & 97.6 & 99.1 & 41.7 & 43.2 & 78.0 & 80.5 & 97.1 \\
MILAN \cite{hou_milan_2022} & ViT-B/16 & 89.9 & 90.8 & 94.2 & 94.9 & 97.4 & 95.5 & 95.5 & 90.8 & 91.8 & 98.1 & 98.2 & 99.5 & 99.5 & 99.9 & 90.3 & 90.5 & 92.6 & 96.0 & 99.0 & 96.3 & 96.4 & 97.1 & 98.5 & 99.2 & 85.0 & 85.0 & 84.6 & 94.4 & 96.5 \\
EVA \cite{fang_eva_2022} & ViT-B/16 & 68.3 & 72.8 & 83.1 & 88.8 & 96.0 & 91.2 & 91.8 & 70.2 & 79.3 & 91.2 & 93.1 & 83.4 & 97.7 & 99.0 & 81.8 & 84.9 & 83.6 & 95.4 & 98.5 & 74.8 & 79.5 & 54.8 & 94.6 & 97.1 & 28.6 & 32.5 & 27.2 & 69.1 & 89.8 \\
PixMIM \cite{liu_pixmim_2023} & ViT-B/16 & 59.4 & 70.5 & 81.7 & 87.0 & 94.7 & 86.2 & 87.6 & 52.0 & 71.5 & 84.9 & 90.6 & 83.8 & 98.0 & 97.7 & 71.1 & 83.0 & 78.7 & 93.2 & 97.2 & 67.3 & 80.4 & 59.1 & 92.9 & 94.3 & 25.7 & 35.8 & 23.4 & 70.9 & 80.0 \\
    \end{tabular}
\end{adjustbox}
    \begin{adjustbox}{width=\textwidth}
    \begin{tabular}{lccccccccccccccccccccccccccccccc}
    \toprule
    & & \multicolumn{30}{c}{Imagenet-D} \\
        & & \multicolumn{5}{c}{Clipart} & \multicolumn{5}{c}{Infograph} & \multicolumn{5}{c}{Painting} & \multicolumn{5}{c}{Quickdraw} & \multicolumn{5}{c}{Real} & \multicolumn{5}{c}{Sketch} \\
    \cmidrule(lr){3-7} \cmidrule(lr){8-12} \cmidrule(lr){13-17} \cmidrule(lr){18-22} \cmidrule(lr){23-27} \cmidrule(lr){28-32}
         Method & Backbone & kNN & kNN\textsuperscript{N} & LP & LP\textsuperscript{BN} & FT & kNN & kNN\textsuperscript{N}  & LP & LP\textsuperscript{BN} & FT & kNN & kNN\textsuperscript{N} & LP & LP\textsuperscript{BN} & FT & kNN & kNN\textsuperscript{N} & LP & LP\textsuperscript{BN} & FT & kNN  & kNN\textsuperscript{N} & LP & LP\textsuperscript{BN} & FT & kNN & kNN\textsuperscript{N} & LP & LP\textsuperscript{BN} & FT \\
\midrule
Jigsaw \cite{noroozi_unsupervised_2016} & RN-50 & 31.4 & 32.3 & 42.1 & 58.2 & 83.7 & 18.2 & 18.3 & 22.6 & 28.1 & 50.9 & 28.4 & 28.1 & 44.9 & 55.4 & 77.5 & 55.0 & 56.9 & 38.4 & 53.0 & 92.5 & 52.9 & 52.9 & 69.0 & 77.0 & 91.2 & 28.7 & 29.6 & 35.9 & 48.3 & 78.1 \\
rotnet \cite{gidaris_unsupervised_2018} & RN-50 & 42.0 & 42.9 & 56.2 & 68.3 & 85.9 & 18.2 & 19.8 & 27.8 & 33.6 & 54.2 & 38.1 & 37.5 & 57.4 & 64.6 & 79.3 & 59.9 & 59.9 & 62.0 & 76.8 & 92.7 & 60.8 & 60.8 & 76.0 & 81.8 & 92.0 & 38.0 & 38.7 & 50.4 & 61.1 & 80.2 \\
npid \cite{wu_unsupervised_2018} & RN-50 & 49.0 & 48.3 & 22.4 & 69.0 & 83.1 & 27.4 & 27.7 & 15.0 & 35.3 & 51.2 & 47.6 & 47.6 & 35.7 & 67.4 & 77.2 & 65.6 & 62.5 & 26.4 & 73.2 & 92.1 & 74.2 & 74.5 & 62.7 & 84.3 & 91.2 & 45.5 & 44.5 & 21.9 & 60.4 & 77.8 \\
Sela-v2 \cite{asano_self-labelling_2020} & RN-50 & 55.9 & 54.5 & 70.8 & 76.2 & 90.0 & 31.2 & 32.9 & 40.7 & 44.0 & 62.8 & 61.0 & 61.2 & 77.4 & 78.9 & 87.3 & 55.1 & 49.1 & 60.4 & 71.3 & 92.2 & 86.0 & 86.2 & 91.3 & 91.6 & 94.8 & 51.4 & 51.2 & 65.8 & 69.8 & 84.9 \\
npid++ \cite{wu_unsupervised_2018, misra_self-supervised_2019} & RN-50 & 42.7 & 40.7 & 64.6 & 68.3 & 86.4 & 27.2 & 27.5 & 34.6 & 35.0 & 54.4 & 45.8 & 45.3 & 67.4 & 66.6 & 80.1 & 48.4 & 37.8 & 56.8 & 71.6 & 92.5 & 74.3 & 74.3 & 86.8 & 85.5 & 92.3 & 35.8 & 35.3 & 51.2 & 55.8 & 80.5 \\
PIRL \cite{misra_self-supervised_2019} & RN-50 & 56.2 & 55.1 & 70.9 & 76.5 & 87.5 & 30.3 & 32.1 & 38.0 & 41.8 & 57.0 & 56.9 & 56.6 & 72.5 & 74.0 & 82.0 & 57.1 & 45.2 & 59.5 & 68.2 & 92.5 & 81.7 & 81.9 & 88.3 & 88.7 & 92.8 & 52.4 & 51.3 & 63.6 & 68.4 & 81.8 \\
clusterfit \cite{yan_clusterfit_2019} & RN-50 & 55.2 & 56.9 & 72.6 & 73.4 & 85.5 & 27.2 & 28.8 & 36.6 & 37.0 & 52.8 & 53.3 & 54.1 & 68.6 & 69.7 & 79.7 & 66.7 & 67.0 & 78.5 & 78.9 & 91.5 & 76.8 & 77.3 & 85.5 & 85.9 & 91.9 & 50.0 & 51.0 & 64.5 & 65.1 & 78.9 \\
Deepcluster-v2 \cite{caron_deep_2018,caron_unsupervised_2020} & RN-50 & 61.5 & 62.2 & 79.9 & 80.3 & 90.5 & 35.3 & 37.1 & 47.8 & 47.4 & 62.4 & 64.7 & 65.3 & 83.0 & 82.1 & 87.7 & 57.5 & 58.0 & 72.1 & 75.8 & 92.4 & 88.5 & 88.7 & 93.6 & 92.7 & 94.9 & 56.1 & 56.5 & 73.3 & 73.3 & 85.5 \\
SwAV \cite{caron_unsupervised_2020} & RN-50 & 59.7 & 60.7 & 78.4 & 79.5 & 90.2 & 35.1 & 36.9 & 46.5 & 47.1 & 62.7 & 64.4 & 65.0 & 82.1 & 81.6 & 87.5 & 55.9 & 54.1 & 73.3 & 78.2 & 92.4 & 88.2 & 88.5 & 93.2 & 92.3 & 94.9 & 55.0 & 55.9 & 72.0 & 73.3 & 85.6 \\
SimCLR \cite{chen_simple_2020} & RN-50 & 54.6 & 54.8 & 61.9 & 77.2 & 89.8 & 30.6 & 33.3 & 32.3 & 42.3 & 60.8 & 60.4 & 59.8 & 72.3 & 78.5 & 86.2 & 49.5 & 35.4 & 51.9 & 75.0 & 92.5 & 84.5 & 84.3 & 88.6 & 90.8 & 94.2 & 51.9 & 52.1 & 58.1 & 71.2 & 84.2 \\
MoCo v2 \cite{chen_improved_2020} & RN-50 & 58.1 & 60.7 & 33.0 & 81.8 & 87.6 & 31.9 & 35.2 & 18.1 & 48.3 & 57.1 & 64.3 & 64.2 & 53.4 & 81.0 & 82.5 & 57.0 & 53.6 & 35.7 & 76.0 & 92.7 & 86.1 & 86.3 & 80.8 & 92.3 & 93.2 & 55.3 & 56.7 & 31.8 & 75.7 & 81.9 \\
SimSiam \cite{chen_exploring_2020} & RN-50 & 60.5 & 62.2 & 5.3 & 82.6 & 84.5 & 33.3 & 36.3 & 6.3 & 48.5 & 52.4 & 65.1 & 64.9 & 7.3 & 82.0 & 78.7 & 63.5 & 59.8 & 32.4 & 81.8 & 92.5 & 86.7 & 86.8 & 31.0 & 92.3 & 92.0 & 58.5 & 59.2 & 6.0 & 76.7 & 78.5 \\
BYOL \cite{grill_bootstrap_2020} & RN-50 & 61.2 & 62.6 & 78.6 & 82.7 & 90.9 & 33.8 & 36.7 & 43.4 & 47.4 & 62.7 & 65.6 & 65.0 & 81.7 & 82.3 & 87.3 & 61.2 & 54.3 & 66.1 & 75.9 & 92.6 & 87.4 & 87.4 & 92.8 & 92.6 & 94.7 & 59.2 & 59.6 & 73.3 & 76.2 & 85.5 \\
Barlow Twins \cite{zbontar_barlow_2021} & RN-50 & 61.1 & 62.2 & 82.0 & 81.8 & 88.1 & 34.5 & 36.3 & 47.3 & 46.6 & 56.3 & 65.4 & 64.9 & 82.3 & 80.9 & 85.0 & 62.9 & 59.2 & 72.5 & 75.0 & 91.0 & 87.7 & 87.8 & 93.0 & 91.9 & 94.0 & 59.6 & 59.0 & 75.4 & 75.0 & 83.0 \\
DenseCL \cite{wang_dense_2020} & RN-50 & 51.6 & 52.5 & 41.0 & 76.0 & 87.4 & 28.6 & 29.8 & 23.6 & 42.0 & 56.6 & 59.0 & 58.9 & 56.4 & 75.8 & 82.4 & 52.8 & 55.7 & 33.5 & 76.4 & 92.7 & 82.1 & 82.6 & 79.0 & 90.2 & 93.0 & 50.1 & 50.5 & 39.1 & 68.9 & 81.6 \\
DINO \cite{caron_emerging_2021} & RN-50 & 59.8 & 60.3 & 80.5 & 81.0 & 90.1 & 35.3 & 37.4 & 48.1 & 47.6 & 62.7 & 65.0 & 64.6 & 82.9 & 82.2 & 87.3 & 57.9 & 55.7 & 72.1 & 76.6 & 92.2 & 88.5 & 88.4 & 93.6 & 92.8 & 94.9 & 56.1 & 56.4 & 73.4 & 73.8 & 85.2 \\
MoCo v3 \cite{chen_empirical_2021} & RN-50 & 68.0 & 66.6 & 83.7 & 84.0 & 91.3 & 35.1 & 36.7 & 48.1 & 48.5 & 63.7 & 68.1 & 66.7 & 84.1 & 83.4 & 88.1 & 65.6 & 59.9 & 74.8 & 78.6 & 92.8 & 88.3 & 87.8 & 93.6 & 93.0 & 95.0 & 64.4 & 63.1 & 78.2 & 78.1 & 86.1 \\
DINO \cite{caron_emerging_2021} & ViT-B/16 & 73.4 & 73.5 & 85.0 & 85.6 & 92.1 & 47.2 & 47.0 & 56.5 & 57.5 & 65.0 & 75.7 & 75.5 & 85.0 & 85.6 & 89.1 & 64.5 & 65.4 & 82.8 & 83.2 & 92.5 & 93.0 & 92.9 & 94.1 & 94.5 & 95.8 & 68.2 & 68.3 & 77.6 & 78.1 & 86.9 \\
iBOT \cite{zhou_ibot_2021} & ViT-B/16 & 74.6 & 74.4 & 86.5 & 87.0 & 93.4 & 47.8 & 47.1 & 57.4 & 58.5 & 67.1 & 76.1 & 75.8 & 85.8 & 86.5 & 90.7 & 66.7 & 67.6 & 85.2 & 85.5 & 93.1 & 93.1 & 93.1 & 94.5 & 94.8 & 96.3 & 67.6 & 67.1 & 78.0 & 78.7 & 88.6 \\
MoCo v3 \cite{chen_empirical_2021} & ViT-B/16 & 71.1 & 71.5 & 60.9 & 85.3 & 89.5 & 42.7 & 42.9 & 35.9 & 54.2 & 60.6 & 71.8 & 72.3 & 73.8 & 84.3 & 88.7 & 67.4 & 68.2 & 58.7 & 85.0 & 91.1 & 91.3 & 91.4 & 91.4 & 94.1 & 95.9 & 63.8 & 64.3 & 56.1 & 76.2 & 84.8 \\
MAE \cite{he_masked_2021} & ViT-B/16 & 39.2 & 62.9 & 47.3 & 81.0 & 89.7 & 21.0 & 33.7 & 25.2 & 47.4 & 61.2 & 28.7 & 56.4 & 54.0 & 78.2 & 87.0 & 66.6 & 70.8 & 55.5 & 83.1 & 92.2 & 58.9 & 82.0 & 81.9 & 91.4 & 94.8 & 40.8 & 54.8 & 39.2 & 71.2 & 83.8 \\
MaskFeat \cite{wei_masked_2022} & ViT-B/16 & 29.4 & 37.5 & 8.1 & 65.7 & 81.3 & 16.1 & 21.3 & 8.0 & 32.3 & 46.5 & 24.1 & 39.2 & 11.1 & 66.3 & 76.7 & 44.0 & 55.3 & 7.7 & 70.6 & 89.7 & 46.7 & 63.8 & 14.8 & 82.3 & 90.6 & 26.1 & 34.7 & 7.9 & 54.3 & 74.6 \\
BEiT v2 \cite{peng_beit_2022} & ViT-B/16 & 77.4 & 77.1 & 89.3 & 88.8 & 95.1 & 43.0 & 42.6 & 59.8 & 59.0 & 71.8 & 73.7 & 73.4 & 87.6 & 87.0 & 93.3 & 69.9 & 71.0 & 83.1 & 84.7 & 93.5 & 92.2 & 92.0 & 95.2 & 95.2 & 97.3 & 68.8 & 68.7 & 82.1 & 81.7 & 90.6 \\
MILAN \cite{hou_milan_2022} & ViT-B/16 & 84.8 & 85.0 & 85.1 & 90.8 & 94.8 & 54.6 & 54.9 & 55.1 & 65.3 & 73.4 & 77.7 & 77.8 & 85.6 & 88.9 & 93.3 & 74.5 & 75.0 & 67.8 & 85.1 & 92.6 & 94.6 & 94.6 & 95.8 & 96.3 & 97.4 & 76.5 & 76.9 & 76.6 & 84.1 & 90.1 \\
EVA \cite{fang_eva_2022} & ViT-B/16 & 58.5 & 62.1 & 50.3 & 82.4 & 90.4 & 34.6 & 36.3 & 28.6 & 51.0 & 64.3 & 55.3 & 59.1 & 57.9 & 79.8 & 88.8 & 75.5 & 76.5 & 62.2 & 84.6 & 92.2 & 82.8 & 85.3 & 85.7 & 92.8 & 96.0 & 50.0 & 52.9 & 41.4 & 72.7 & 84.9 \\
PixMIM \cite{liu_pixmim_2023} & ViT-B/16 & 50.8 & 63.6 & 47.7 & 81.2 & 86.3 & 28.2 & 34.7 & 25.9 & 48.2 & 57.4 & 45.0 & 59.0 & 56.4 & 78.6 & 84.7 & 67.6 & 70.3 & 54.4 & 82.0 & 91.2 & 74.7 & 83.5 & 83.1 & 91.5 & 94.6 & 44.8 & 54.6 & 39.2 & 71.2 & 80.5 \\
    \bottomrule
    \end{tabular}
\end{adjustbox}
    \begin{adjustbox}{width=.6\textwidth}
    \begin{tabular}{lcccccccccccccccc}
    & & \multicolumn{15}{c}{iNaturalist mini} \\
        & & \multicolumn{5}{c}{Target: Family} & \multicolumn{5}{c}{Target: Genus} & \multicolumn{5}{c}{Target: Species} \\
    \cmidrule(lr){3-7} \cmidrule(lr){8-12} \cmidrule(lr){13-17}
         Method & Backbone & kNN & kNN\textsuperscript{N} & LP & LP\textsuperscript{BN} & FT & kNN & kNN\textsuperscript{N}  & LP & LP\textsuperscript{BN} & FT & kNN & kNN\textsuperscript{N}  & LP & LP\textsuperscript{BN} & FT \\
\midrule
Jigsaw \cite{noroozi_unsupervised_2016} & RN-50 & 25.0 & 25.4 & 28.7 & 41.4 & 89.5 & 6.3 & 6.9 & 12.4 & 17.1 & 66.7 & 4.7 & 5.1 & 9.4 & 12.4 & 62.3 \\
rotnet \cite{gidaris_unsupervised_2018} & RN-50 & 28.1 & 28.3 & 42.0 & 46.6 & 89.6 & 8.2 & 8.2 & 16.5 & 22.7 & 69.8 & 6.5 & 6.6 & 13.6 & 18.0 & 65.9 \\
npid \cite{wu_unsupervised_2018} & RN-50 & 40.2 & 40.4 & 37.5 & 56.6 & 88.8 & 17.6 & 18.4 & 14.9 & 34.6 & 66.7 & 14.1 & 14.4 & 12.6 & 28.3 & 64.0 \\
Sela-v2 \cite{asano_self-labelling_2020} & RN-50 & 55.6 & 57.1 & 66.2 & 69.2 & 91.7 & 30.0 & 33.4 & 42.1 & 52.0 & 81.3 & 23.2 & 25.9 & 34.5 & 43.5 & 76.9 \\
npid++ \cite{wu_unsupervised_2018, misra_self-supervised_2019} & RN-50 & 41.0 & 41.2 & 57.7 & 57.8 & 90.7 & 18.1 & 19.0 & 34.4 & 36.8 & 75.1 & 13.9 & 14.4 & 27.8 & 29.9 & 71.0 \\
PIRL \cite{misra_self-supervised_2019} & RN-50 & 48.8 & 48.6 & 39.5 & 61.6 & 90.8 & 24.3 & 25.0 & 38.4 & 42.1 & 76.1 & 18.4 & 19.1 & 31.3 & 34.4 & 71.6 \\
clusterfit \cite{yan_clusterfit_2019} & RN-50 & 38.5 & 39.0 & 54.2 & 54.4 & 89.2 & 14.9 & 15.9 & 28.0 & 28.8 & 67.2 & 11.1 & 11.7 & 21.4 & 22.2 & 61.4 \\
Deepcluster-v2 \cite{caron_deep_2018,caron_unsupervised_2020} & RN-50 & 62.3 & 64.2 & 74.7 & 75.0 & 92.2 & 37.2 & 41.4 & 58.8 & 61.8 & 82.0 & 29.7 & 33.6 & 51.2 & 54.1 & 77.5 \\
SwAV \cite{caron_unsupervised_2020} & RN-50 & 60.0 & 62.5 & 72.9 & 73.8 & 92.2 & 33.9 & 39.3 & 52.4 & 58.2 & 81.6 & 26.3 & 31.1 & 44.5 & 50.4 & 77.0 \\
SimCLR \cite{chen_simple_2020} & RN-50 & 52.1 & 51.8 & 59.9 & 67.0 & 90.8 & 27.7 & 29.2 & 34.0 & 47.4 & 78.1 & 21.2 & 22.3 & 27.1 & 38.8 & 72.8 \\
MoCo v2 \cite{chen_improved_2020} & RN-50 & 56.7 & 57.4 & 52.1 & 73.7 & 91.9 & 33.5 & 34.7 & 25.7 & 58.6 & 77.4 & 26.0 & 27.7 & 20.3 & 50.1 & 73.9 \\
SimSiam \cite{chen_exploring_2020} & RN-50 & 55.9 & 56.5 & 20.8 & 71.6 & 91.0 & 32.1 & 34.4 & 20.5 & 55.9 & 69.5 & 24.8 & 26.8 & 15.8 & 48.1 & 67.9 \\
BYOL \cite{grill_bootstrap_2020} & RN-50 & 60.8 & 62.4 & 70.6 & 73.3 & 91.6 & 38.5 & 43.8 & 50.9 & 61.3 & 81.5 & 30.5 & 35.7 & 43.0 & 52.9 & 77.1 \\
Barlow Twins \cite{zbontar_barlow_2021} & RN-50 & 60.8 & 62.0 & 72.9 & 72.9 & 87.1 & 38.0 & 41.4 & 56.3 & 58.4 & 73.2 & 29.8 & 33.8 & 48.9 & 51.1 & 65.7 \\
DenseCL \cite{wang_dense_2020} & RN-50 & 48.8 & 49.5 & 49.2 & 65.5 & 91.5 & 25.5 & 26.9 & 21.0 & 46.6 & 75.2 & 19.0 & 19.8 & 16.4 & 38.4 & 72.7 \\
DINO \cite{caron_emerging_2021} & RN-50 & 63.8 & 64.0 & 75.4 & 76.2 & 92.2 & 40.0 & 41.3 & 55.7 & 61.1 & 81.4 & 31.5 & 33.0 & 47.2 & 52.3 & 76.5 \\
MoCo v3 \cite{chen_empirical_2021} & RN-50 & 63.8 & 62.7 & 72.5 & 73.1 & 92.0 & 44.8 & 46.3 & 58.3 & 62.7 & 82.6 & 37.3 & 39.4 & 51.2 & 55.2 & 78.7 \\
DINO \cite{caron_emerging_2021} & ViT-B/16 & 78.8 & 79.1 & 83.2 & 83.7 & 94.3 & 68.5 & 68.6 & 80.0 & 81.2 & 87.0 & 59.7 & 60.1 & 74.3 & 75.8 & 83.3 \\
iBOT \cite{zhou_ibot_2021} & ViT-B/16 & 77.5 & 77.9 & 83.5 & 84.0 & 95.1 & 66.6 & 66.9 & 80.4 & 81.5 & 88.5 & 58.3 & 58.5 & 74.2 & 75.8 & 85.3 \\
MoCo v3 \cite{chen_empirical_2021} & ViT-B/16 & 70.3 & 70.9 & 69.5 & 80.8 & 92.2 & 52.0 & 53.0 & 48.4 & 74.6 & 85.4 & 42.4 & 43.9 & 41.2 & 67.2 & 81.4 \\
MAE \cite{he_masked_2021} & ViT-B/16 & 29.4 & 47.3 & 55.7 & 71.3 & 93.3 & 8.4 & 22.9 & 20.1 & 55.8 & 84.3 & 6.6 & 17.9 & 16.4 & 48.3 & 80.2 \\
MaskFeat \cite{wei_masked_2022} & ViT-B/16 & 22.1 & 35.2 & 20.7 & 57.4 & 90.0 & 5.5 & 12.8 & 2.2 & 32.3 & 70.5 & 4.7 & 9.7 & 2.0 & 25.4 & 67.5 \\
BEiT v2 \cite{peng_beit_2022} & ViT-B/16 & 67.3 & 68.2 & 83.5 & 83.8 & 96.3 & 36.2 & 37.1 & 69.3 & 69.1 & 91.8 & 27.2 & 27.9 & 61.1 & 60.8 & 89.3 \\
MILAN \cite{hou_milan_2022} & ViT-B/16 & 78.2 & 78.6 & 78.3 & 85.5 & 96.0 & 60.7 & 61.4 & 62.1 & 79.0 & 90.5 & 51.7 & 52.3 & 55.3 & 73.3 & 87.1 \\
EVA \cite{fang_eva_2022} & ViT-B/16 & 50.4 & 54.4 & 60.8 & 76.2 & 95.0 & 20.9 & 24.4 & 22.1 & 54.6 & 87.5 & 15.4 & 18.2 & 17.8 & 45.8 & 83.8 \\
PixMIM \cite{liu_pixmim_2023} & ViT-B/16 & 40.1 & 49.6 & 56.8 & 71.5 & 92.6 & 15.4 & 23.7 & 21.9 & 55.4 & 82.0 & 11.0 & 17.9 & 18.1 & 47.1 & 78.3 \\
\bottomrule
\end{tabular}
\end{adjustbox}
    \end{table*}

\clearpage
\section{Additional Visualizations}

\begin{figure*}[ht!]
    \centering
    \includegraphics[width=\textwidth]{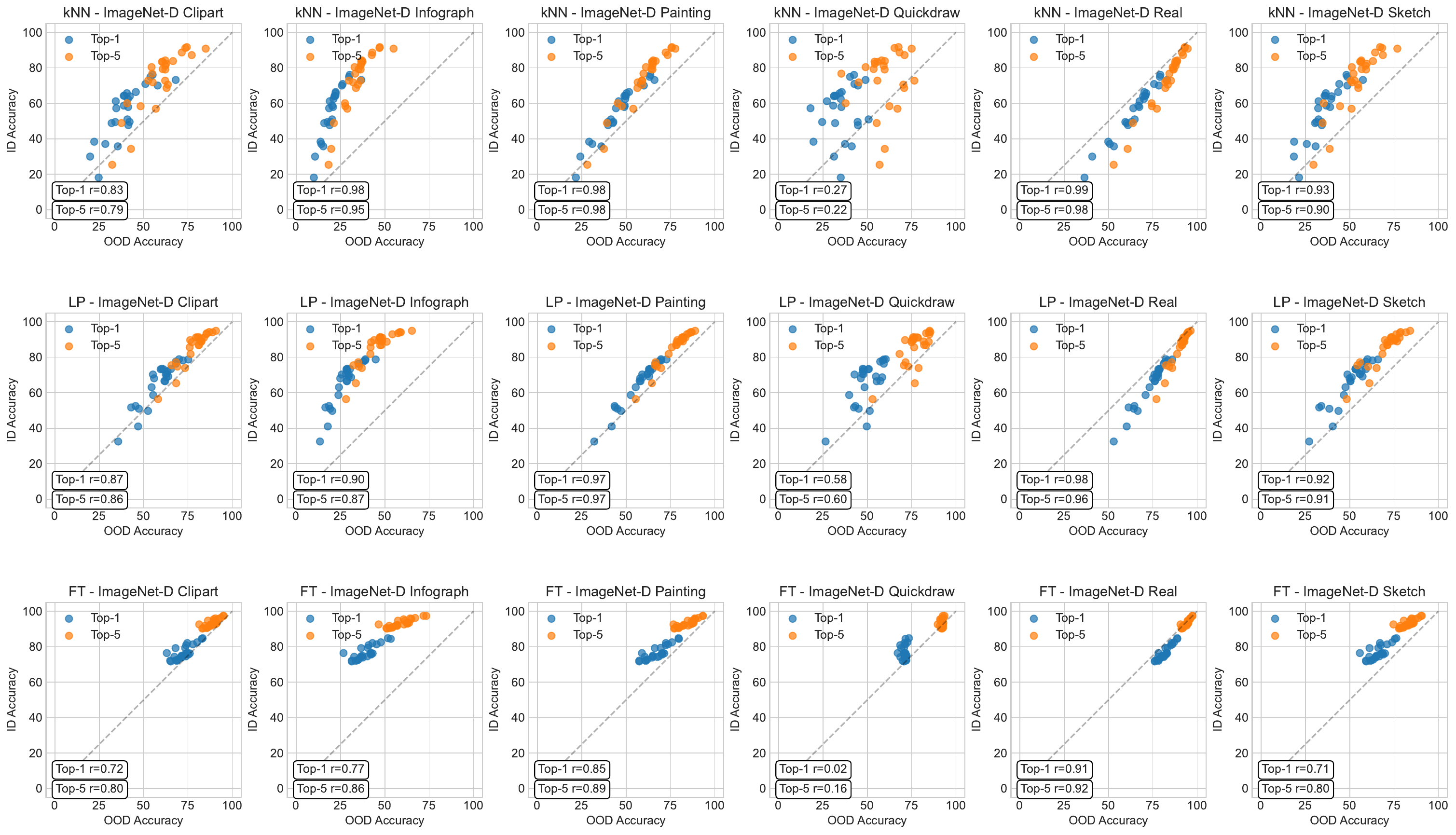}
    \caption{ImageNet-D ID vs. OOD accuracies on different protocols. We compare both top-1 and top-5 classification accuracies. Correlation coefficients $r$ are calculated using Spearman's rank correlation.}
    \label{fig:imagenet-d-panels}
\end{figure*}

\begin{figure}[ht!]
\input{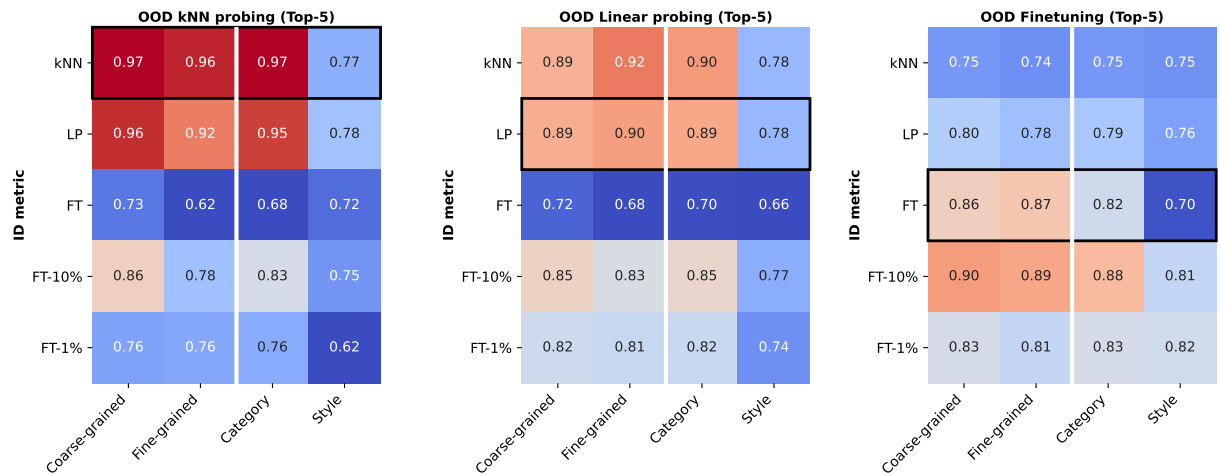}
\caption{Spearman rank correlations of \textit{top-5} classification accuracies derived from in-domain and out-of-domain protocols under certain types of domain shift. See \Cref{fig:domainshift} for more details.
}
\label{fig:domainshift-top5}
\end{figure}

\begin{figure}[ht]
\centering

\begin{subfigure}[t]{0.45\textwidth} 
    \centering
        \includegraphics[width=\textwidth]{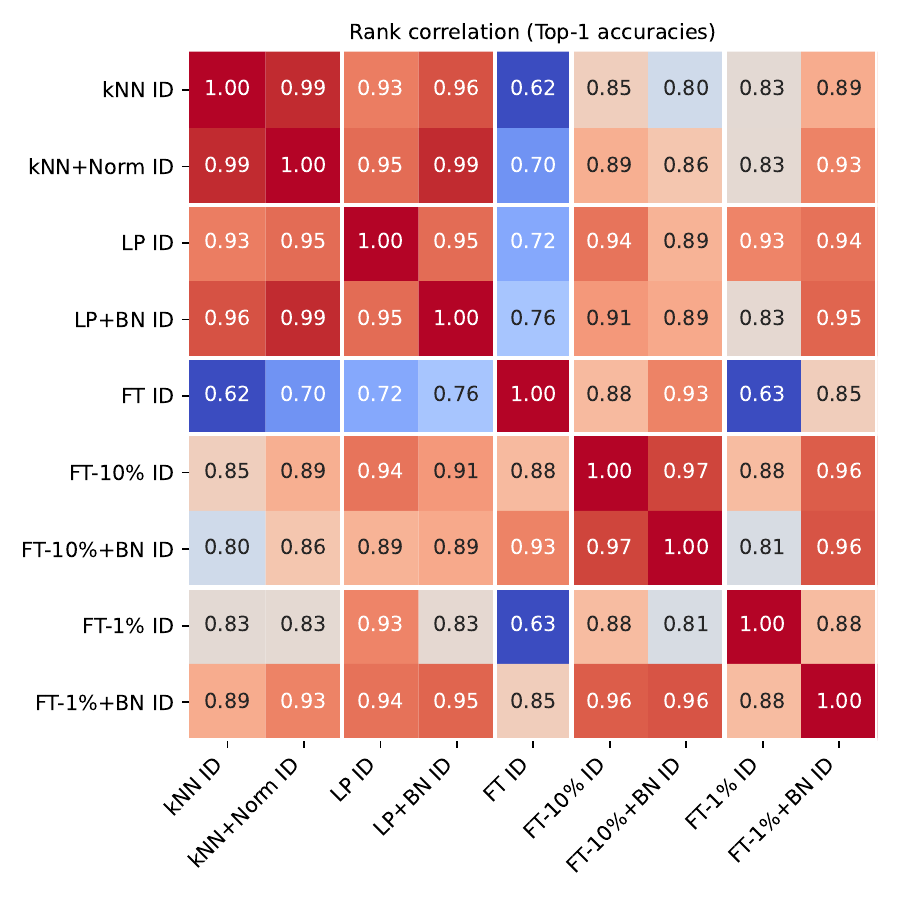}
\end{subfigure}
\hfill
\begin{subfigure}[t]{0.45\textwidth}
    \centering
        \includegraphics[width=\textwidth]{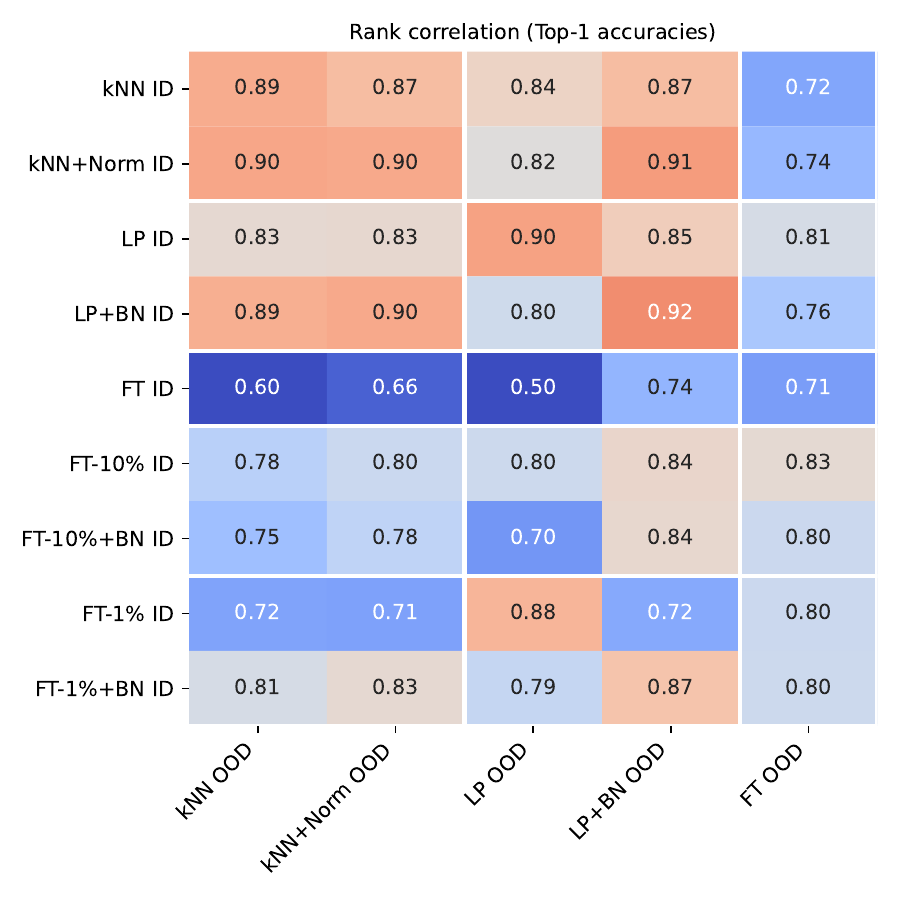}

\end{subfigure}

\caption{Extended version of \Cref{fig:corr-matrix}. We show both versions, with and without feature normalization for all protocols but 100\% fine-tuning.
}
\label{fig:corr-matrix-extended}
\end{figure}

\section{Fine-grained experiments}\label{appendix:inat}

The iNaturalist data set comes with different levels of hierarchical classes. Taxonomic closeness is a rough proxy for the degree of visual similarity and implies different degrees of fine-graininess of visual features \citep{cole_when_2022}. We use the three most fine-grained targets, ``Family'', ``Genus'', and ``Species'' to estimate whether evaluation protocols or models are more or less sensitive to fine-grained features than others.
For ``Family'' we use the full iNat mini data set comprising 1103 classes. By definition, more fine-grained classes would increase the number of classes when the whole dataset is used. To ensure a fair comparison with ``Genus'' and ``Species'' targets, we create subsampled datasets that have the same number of classes. At the same time, we ensure that the number of categories higher-level categories for a subset is as small as possible. In detail, we realize this as follows:
For the ``Genus'' subset, we select the 256 families from the ``Chordata'' phylum that contain the most species, resulting in 1103 ``Genus'' categories.
For the ``Species'' subset, we sort the previously defined Genus subset by the number of species and pick the top 277 categories, resulting in 1103 ``Species'' targets.

\begin{figure}[ht!]
\input{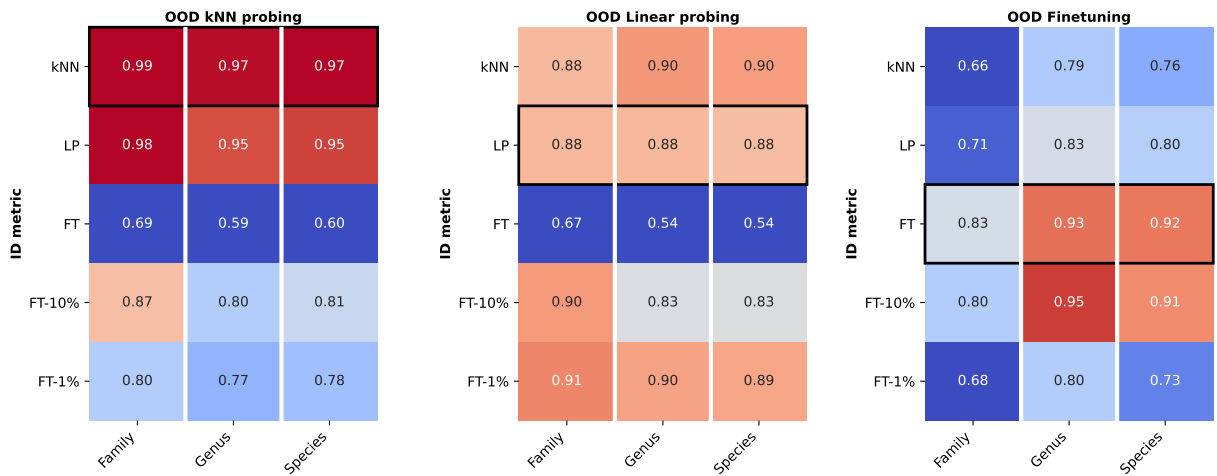}
\caption{Spearman rank correlations of top-1 classification accuracies for three different targets in the iNaturalist mini dataset.
}
\label{fig:domainshift-inat}
\end{figure}

\begin{figure}[ht!]
    \centering
\includegraphics[width=.6\textwidth]{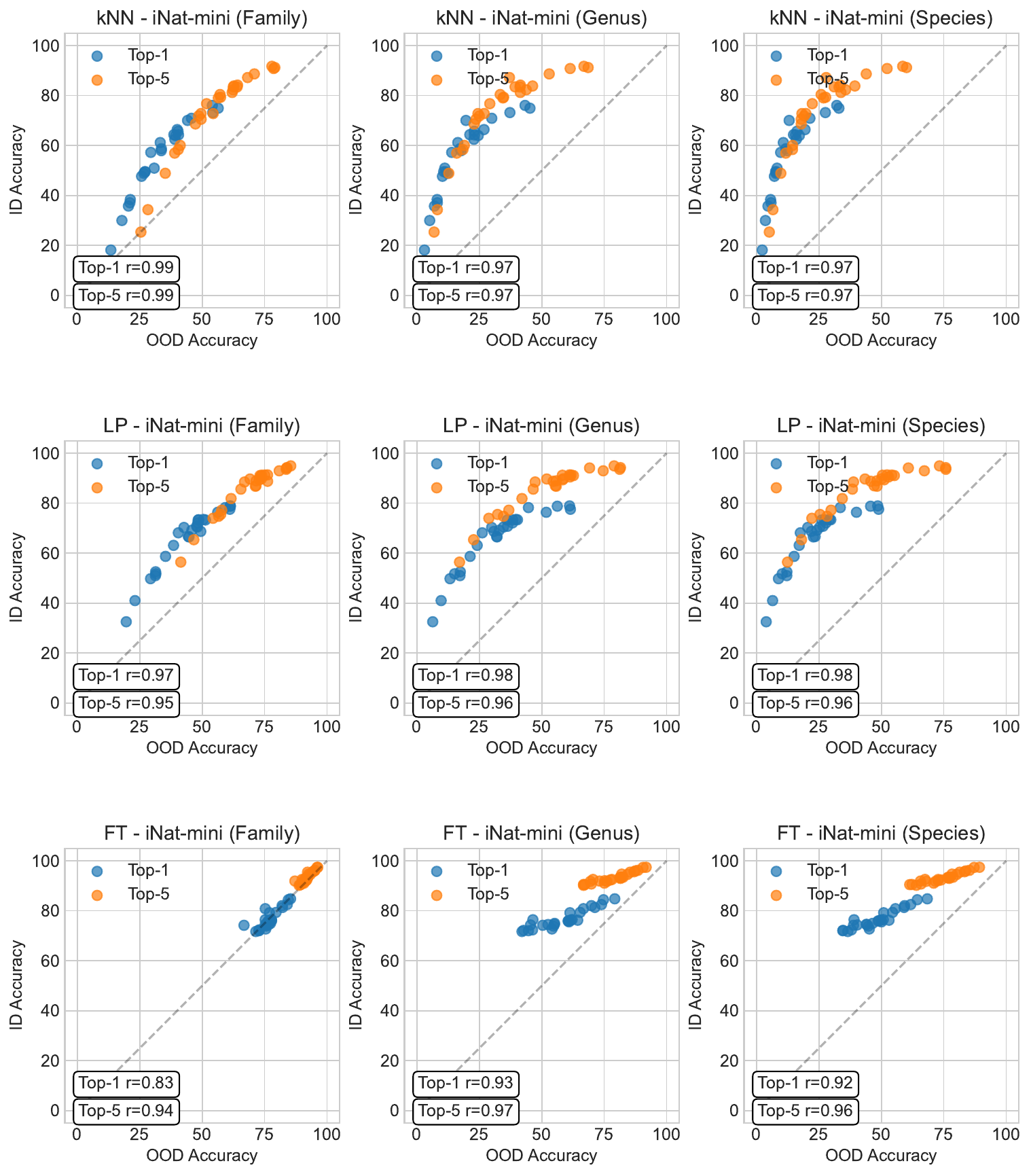}
    \caption{iNaturalist ID vs. OOD accuracies with different targets (family, genus, species) on different protocols. We compare both top-1 and top-5 classification accuracies. Correlation coefficients $r$ are calculated using Spearman's rank correlation.}
    \label{fig:enter-label}
\end{figure}

\newpage
\FloatBarrier
\section{Implementation Details}
\label{sec:implementation_details}

All experiments were run using models implemented in PyTorch \cite{paszke_pytorch_2019} using the \texttt{timm} \cite{wightman_pytorch_2019} and \texttt{mmselfsup} \cite{mmselfsup_contributors_mmselfsup_2021} libraries. All end-to-end fine-tuning and linear probing were trained using a Stochastic Gradient Descent optimizer with 100 total epochs (5 warmup epochs followed by Cosine Annealing decay) and a base learning rate of 0.1. Few-shot fine-tuning on ImageNet was trained on 30 epochs only as proposed by \cite{grill_bootstrap_2020}. We found this few-shot setup to be the most stable compared to others that have different learning rates for head and backbone (see, e.g., \cite{caron_unsupervised_2020, zbontar_barlow_2021}). 
The effective batch sizes varied per data set and can be found in \Cref{tab:implementation_details}.
When training vision transformers, we also use layerwise learning rate decay (0.65), label smoothing (0.1), and drop path (0.2), inspired by \cite{he_masked_2021}.

\begin{table}[ht!]
    \centering
    \caption{Chosen effective batch sizes for different data sets used in this study.}
    \begin{tabular}{cccc}
\toprule
Dataset & End-to-end fine-tuning & linear probe \\
\midrule
ImageNet-1k (full) & 256 & 1024 \\
ImageNet-1k (10\%) & 256 & - \\
ImageNet-1k (1\%) & 256 & - \\
Pascal VOC & 32 & 32 \\
Caltech-256 & 32 & 32 \\
CUB & 32 & 32 \\
Cifar100 & 128 & 128 \\
iNaturalist mini, Family & 256 & 1024 \\
iNaturalist mini, Genus & 256 & 1024 \\
iNaturalist mini, Species & 256 & 1024 \\
ImageNet-D Clipart & 128 & 128 \\
ImageNet-D Infograph & 128 & 128 \\
ImageNet-D Painting & 128 & 128 \\
ImageNet-D Quickdraw & 256 & 256 \\
ImageNet-D Real & 256 & 256 \\
ImageNet-D Sketch & 128 & 128 \\
\bottomrule
\end{tabular}

    \label{tab:implementation_details}
\end{table}

KNN-probing was implemented with \texttt{scikit-learn} \cite{pedregosa_scikit-learn_2011} using Euclidean distance, a brute force solver, and $k=20$ neighbors (as proposed by \cite{caron_emerging_2021}).

For ResNet-50, we use the last 2048-dimensional (pre-logit) representation to report probing accuracy. \cite{kolesnikov_revisiting_2019} described that this consistently works better for this architecture than using any intermediate latent representation.
For Vision Transformers, we follow the protocol of most implementations by training the linear classifier on the 768-dimensional \texttt{cls}-token.

Certain models---especially those using masked image modeling --- achieve a significantly better linear probe accuracy when batch normalization is used between the model outputs and the linear classifier, as suggested by \cite{he_masked_2021} and verified by \cite{lee_rethinking_2023}. Based on our own experiments, we can see that normalization has either a strong positive or an insignificant effect on the accuracy depending on the model (see \cref{tab:results_all}). Accordingly, we decided to use batch normalization for linear probing and also normalize latent representations before kNN probing.
\Cref{tab:checkpoints} lists the sources of the pre-trained model checkpoints we used.

\begin{table}[ht!]
    \centering
    \small
    \caption{
    Sources of the pre-trained model checkpoints used in this study.
    }
    \label{tab:checkpoints}
    \begin{tabular}{lll}
    \toprule
         Method & Backbone~ & Source \\
\midrule
Jigsaw & RN-50  & \href{https://github.com/facebookresearch/vissl/blob/main/MODEL_ZOO.md)}{\texttt{VISSL model zoo}}          \\
rotnet & RN-50  &    \href{https://github.com/facebookresearch/vissl/blob/main/MODEL_ZOO.md)}{\texttt{VISSL model zoo}}        \\
npid & RN-50  &  \href{https://github.com/facebookresearch/vissl/blob/main/MODEL_ZOO.md)}{\texttt{VISSL model zoo}}         \\
Sela-v2 & RN-50  &  \href{https://github.com/facebookresearch/swav}{\texttt{SwAV official Github}}      \\
npid++ & RN-50  &  \href{https://github.com/facebookresearch/vissl/blob/main/MODEL_ZOO.md)}{\texttt{VISSL model zoo}} \\
PIRL & RN-50  &   \href{https://github.com/facebookresearch/vissl/blob/main/MODEL_ZOO.md)}{\texttt{VISSL model zoo}}       \\
clusterfit & RN-50  &    \href{https://github.com/facebookresearch/vissl/blob/main/MODEL_ZOO.md)}{\texttt{VISSL model zoo}}       \\
Deepcluster-v2 & RN-50  &    \href{https://dl.fbaipublicfiles.com/deepcluster/deepclusterv2_800ep_pretrain.pth.tar}{\texttt{Deepcluster official Github}}       \\
SwAV & RN-50  &    \href{https://github.com/facebookresearch/swav}{\texttt{SwAV official Github}}       \\
SimCLR & RN-50  &   \href{https://github.com/facebookresearch/vissl/blob/main/MODEL_ZOO.md)}{\texttt{VISSL model zoo}}        \\
MoCo v2 & RN-50 &    \href{https://github.com/facebookresearch/moco}{\texttt{MoCoV2 official Github}}       \\
SimSiam & RN-50 &    \href{https://mmselfsup.readthedocs.io/en/dev-1.x/model_zoo.html}{\texttt{MMSelfSup model zoo}}      \\
BYOL & RN-50 &    \href{https://github.com/yaox12/BYOL-PyTorch}{\texttt{Github (inofficial)}}       \\
Barlow Twins & RN-50 &    \href{https://mmselfsup.readthedocs.io/en/dev-1.x/model_zoo.html}{\texttt{MMSelfSup model zoo}}     \\
DenseCL & RN-50 &    \href{https://github.com/WXinlong/DenseCL}{\texttt{DenseCL official Github}}       \\
DINO & RN-50 &   \href{https://github.com/facebookresearch/dino}{\texttt{DINO official Github}}       \\
MoCo v3 & RN-50 &   \href{https://github.com/facebookresearch/moco-v3}{\texttt{MoCoV3 official Github}}        \\
DINO & ViT-B/16 &   \href{https://github.com/facebookresearch/dino}{\texttt{DINO official Github}}        \\
iBOT & ViT-B/16 &  \href{https://github.com/bytedance/ibot}{\texttt{iBOT official Github}}        \\
MoCo v3  & ViT-B/16 & \href{https://github.com/facebookresearch/moco-v3}{\texttt{MoCoV3 official Github}} \\
MAE & ViT-B/16 &   \href{https://github.com/facebookresearch/mae}{\texttt{MAE official Github}}      \\
MaskFeat & ViT-B/16 & \href{https://mmselfsup.readthedocs.io/en/dev-1.x/model_zoo.html}{\texttt{MMSelfSup model zoo}} \\
BEiT v2 & ViT-B/16 & \href{https://github.com/microsoft/unilm/tree/master/beit2}{\texttt{BEiT official Github}}  \\
MILAN & ViT-B/16 & \href{https://github.com/zejiangh/milan}{\texttt{MILAN official Github}} \\
EVA & ViT-B/16 & \href{https://mmselfsup.readthedocs.io/en/dev-1.x/model_zoo.html}{\texttt{MMSelfSup model zoo}} \\
PixMIM & ViT-B/16 & \href{https://mmselfsup.readthedocs.io/en/dev-1.x/model_zoo.html}{\texttt{MMSelfSup model zoo}} \\
    \bottomrule
    \end{tabular}
    \end{table}

\begin{table}[ht!]
\caption{Overview of how datasets were categorized in our domain shift experiments (\cref{fig:domainshift}). We define domain shifts w.r.t. ImageNet-1k.}
\label{tab:dataset-assignment}
\centering
\begin{tabular}{lcccc}
\toprule
Dataset              & ~Coarse~ & ~Fine~ & ~Category~ & ~Style~ \\
\midrule
Pascal VOC           & \checkmark      &      &          &       \\
Caltech256           & \checkmark      &      & \checkmark        &       \\
CUB                  &        & \checkmark    & \checkmark        &       \\
iNat mini Family     & \checkmark      &      & \checkmark        &       \\
iNat mini Genus      &        & \checkmark    &          &       \\
iNat mini Species    &        & \checkmark    &          &       \\
ImageNet-D Clipart   &        &      &          & \checkmark     \\
ImageNet-D Infograph &        &      &          & \checkmark     \\
ImageNet-D Painting  &        &      &          & \checkmark     \\
ImageNet-D Quickdraw &        &      &          & \checkmark     \\
ImageNet-D Sketch    &        &      &          & \checkmark     \\
\bottomrule
\end{tabular}
\end{table}

\clearpage
\section{Approximating Errors}
\label{sec:appendix-errors}

Running the full set of experiments multiple times is costly. Therefore, we randomly selected one model per metric per dataset and ran the same experiment three times in order to approximate a representative error value (see \Cref{tab:errors}). We can see that errors are generally small and conclude that our metrics generated by a single run can be trusted.

\begin{table}[ht!]
    \centering
    \caption{Top-1 accuracies for three different seeds of different models on different datasets. Generally, metrics generated by evaluation protocols are highly reproducible. Note that we do not cover the uncertainty introduced by different pre-training setups.}
    \begin{tabular}{ccccccccc}
    \toprule
    Protocol & Dataset & Method & Backbone & Run~1 & Run~2 & Run~3 & mean & std \\
    \midrule
    Linear Probing & ImageNet & SimCLR & RN-50 & 66.87 & 66.82 & 66.85 & 66.85 & 0.02 \\
    Fine-tuning & ImageNet & DINO & RN-50 & 76.00 & 76.01 & 76.27 & 76.09 & 0.12 \\
    10\% Fine-tuning & ImageNet & MILAN & ViT-B/16 & 78.92 & 78.91 & 78.91 & 78.91 & 0.01 \\
    1\% Fine-tuning & ImageNet & EVA & ViT-B/16 & 41.10 & 46.65 & 46.65 & 44.80 & 2.62 \\
    \midrule
    Linear Probe & Pascal~VOC & Deepcluster~v2 & RN-50 & 85.66 & 85.72 & 85.86 & 85.75 & 0.08 \\
    Fine-tuning & Pascal~VOC & Jigsaw & RN-50 & 64.62 & 64.73 & 63.22 & 64.19 & 0.69 \\
\midrule
    Linear Probe & Caltech-256 & BEiT~v2 & ViT-B/16 & 90.22 & 90.25 & 90.26 & 90.24 & 0.02 \\
    Fine-tuning & Caltech-256 & MoCo~v3 & RN-50 & 88.52 & 88.53 & 88.40 & 88.48 & 0.06 \\
\midrule
    Linear Probe & CUB & DINO & ViT-B/16 & 78.17 & 78.41 & 78.48 & 78.35 & 0.13 \\
    Fine-tuning & CUB & SeLa~v2 & RN-50 & 68.23 & 68.93 & 68.16 & 68.44 & 0.35 \\
\midrule
    Linear Probe & iNat mini (family) & SwAV & RN-50 & 46.94 & 47.27 & 47.13 & 47.11 & 0.14 \\
    Fine-tuning & iNat mini (family) & Barlowtwins & RN-50 & 66.72 & 66.71 & 66.86 & 66.76 & 0.07 \\

    \bottomrule
    \end{tabular}

    \label{tab:errors}
\end{table}

\FloatBarrier
\section{Computational Costs of Protocols}
\begin{table}[htp]
    \centering
    \caption{Estimated computational cost of different protocols.
    GPU-based metrics were estimated based on training on 2x NVIDIA RTX A5000. For k-NN, we neglect CPU-based time required by the classifier and only account for GPU-time used for model inference. The values shown are for ImageNet-1k (in-domain) protocols. Hyperparameters for other datasets can be found in \Cref{sec:implementation_details}.
    }
    \label{tab:computational_cost}
    \begin{tabular}{lcccc}
    \toprule
               & & & \multicolumn{2}{c}{GPU hours} \\
               \cmidrule(lr){4-5}
               Protocol & Batch size & epochs & ResNet-50            & ViT-B/16            \\
    \midrule
kNN      & 1024 & - & 0.6 & 1.3 \\
LP    & 1024 & 100 & 36 & 70  \\
FT       & 256 & 100  & 94 & 184  \\
10\%-FT & 128 & 30  & 3.5 & 6.5  \\
1\%-FT  & 128 & 30  & 1.0 & 1.5  \\
\bottomrule
\end{tabular}
\end{table}

\end{document}